\documentclass[%
 10pt,
 amsmath,amssymb,
 aps, physrev,
  author-year,
 twocolumn, 
 floatfix,
]{revtex4-2}

\usepackage[utf8]{inputenc}
\usepackage[T1]{fontenc}    
\usepackage{CJKutf8}
\usepackage{graphicx}
\usepackage{dcolumn}
\usepackage{bm}
\usepackage{hyperref}
\usepackage{braket}
\hypersetup{
    colorlinks=true,
    linkcolor=blue,
    filecolor=magenta,
    urlcolor=cyan,
    citecolor=blue,
}
\usepackage[mathlines]{lineno}
\newcommand{\mo}{\text{\begin{CJK}{UTF8}{min}も\end{CJK}}}

\begin{document}

\preprint{APS/QNLP-BellTest-Draft}

\title{A quantum semantic framework for natural language processing}

\author{Christopher Agostino}
\email{info@npcworldwi.de}
\homepage{http://www.cjagostino.com}
\affiliation{NPC Worldwide, Bloomington, Indiana 47403, USA}

\author{Quan Le Thien}
\affiliation{Department of Physics, Indiana University, Bloomington, Indiana 47405, USA}
\affiliation{Quantum Science and Engineering Center (QSEC), Indiana University, Bloomington, Indiana 47405, USA}

\author{Molly Apsel}
\affiliation{Cognitive Science Program, Indiana University, Bloomington, Indiana 47405, USA}
\affiliation{Department of Psychological and Brain Sciences, Indiana University, Bloomington, Indiana 47405, USA}

\author{Denizhan Pak}
\affiliation{Cognitive Science Program, Indiana University, Bloomington, Indiana 47405, USA}
\affiliation{Luddy School of Informatics, Computing, and Engineering, Indiana University, Bloomington, Indiana 47405, USA}

\author{Elina Lesyk}
\affiliation{Independent Consultant, Munich 81549, Germany}
\email{info@elinalesyk.com} 

\author{Ashabari Majumdar}
\affiliation{Department of Physics, University of Notre Dame, Notre Dame, IN 46556, USA}
\affiliation{NPC Worldwide, Bloomington, Indiana 47403, USA}

\collaboration{Spontaneity Collaboration}


\begin{abstract}

Semantic degeneracy represents a fundamental property of natural language that extends beyond simple polysemy to encompass the combinatorial explosion of potential interpretations that emerges as semantic expressions increase in complexity. In this work, we argue this property imposes fundamental limitations on Large Language Models (LLMs) and other modern NLP systems, precisely because they operate within natural language itself. Using Kolmogorov complexity, we demonstrate that as an expression's complexity grows, the amount of contextual information required to reliably resolve its ambiguity explodes combinatorially. The computational intractability of recovering a single intended meaning for complex or ambiguous text therefore suggests that the classical view that linguistic forms possess intrinsic meaning in and of themselves is conceptually inadequate. We argue instead that meaning is dynamically actualized through an observer-dependent interpretive act, a process whose non-deterministic nature is most appropriately described by a non-classical, quantum-like logic. To test this hypothesis, we conducted a semantic Bell inequality test using diverse LLM agents. Our experiments yielded average CHSH expectation values from 1.2 to 2.8, with several runs producing values (e.g., 2.3-2.4) in significant violation of the classical boundary ($|S|\leq2$), demonstrating that linguistic interpretation under ambiguity can exhibit non-classical contextuality, consistent with results from human cognition experiments. These results inherently imply that classical frequentist-based analytical approaches for natural language are necessarily lossy. Instead, we propose that Bayesian-style repeated sampling approaches can provide more practically useful and appropriate characterizations of linguistic meaning in context.

\end{abstract}


\maketitle

\section{Introduction}
Prior to the deep-learning revolution of the 2010s, there was a prolonged AI winter spanning from the mid-1980s to the early 2000s. During this period, most researchers in natural language processing and artificial intelligence contended---based on the promises of AI that were unmet in the 1960s and 1970s---that computers on their own would never be capable of producing any kind of general intelligence or simulating the phenomenon that is described as natural language. AI researchers that tried often found themselves drowning in heuristics that broke down with edge cases, usually abandoning the tasks for more readily-solvable problems. See Dreyfus' \textit{What Computers Still Can't Do} \citep{dreyfus1992} to better understand the logical and semantic limitations of artificial reason perceived by researchers prior to the successes of tools making extensive use of large neural networks like Google Translate and others that followed (e.g., transformers, large language models).
Such large neural networks that have now become commonplace for consumer products are often termed Distributional Semantic Models (DSMs) \citep{Harris1954, deerwester1990indexing, LandauerDumais1997, GriffithsEtAl2007, mikolov2013distributed}. DSMs, in effect, infer relevance and meaning from statistical co-occurrences, and they have enabled a variety of highly practical natural language processing (NLP) applications including topic models, sentiment classification, and large language models \citep{DevlinEtAl2018, RadfordEtAl2019gpt2, brown2020language, vaswani2017attention}. By construction, DSMs like BERT-style topic models presuppose that documents possess singular, intrinsic semantic compositions \citep[e.g., early bag-of-words models; see][for contrast with dynamic views]{Harris1954}. 

Because of the practical successes of DSMs in many domains, some researchers have forgotten or not considered many of the initial criticisms of artificial reason that to this day remain sound, instead assuming that these limitations too will be solved by more compute or more data. However, consider the following salient problems in the current field of Natural Language Processing:
\begin{enumerate}
\item
DSM-powered approaches still exhibit limitations particularly when dealing with complex, ambiguous, or context-rich texts \citep{Jones2018, HoffmanEtAl2013}.
\item
The apparent lack of much progress seen in the latest generation of frontier large language models (LLMs)---compared to the leaps seen in prior model releases---appears to have revealed a fundamental barrier in the reliability of semantic problem solving that more compute/data have not alleviated
 (e.g., GPT-4.5, Llama 4, Gemini 2.5, Claude 3.7 Sonnet). Likewise, many frontier researchers have shifted away from prioritizing progress in LLMs in favor of more dynamically capable models (e.g. JEPA \citep{Assran_2023_CVPR}, MAMBA \citep{gu_mamba_2023}, CTM \citep{darlow_ctm_2025}, or modifications to LLMs that allow them to self-improve \citep{transformer2, simonds2025}).
\end{enumerate}

These issues and the evolving research landscape suggest that the challenges faced by DSM-based approaches may stem from a more fundamental issue in linguistics: the problem of semantic degeneracy. This concept, extending beyond simple polysemy, refers to the inherent multiplicity of potential interpretations that arise when processing complex linguistic expressions \citep{TabossiEtAl1987, BinderRayner1998, RaynerEtAl2006}. 

Indeed, the concept intuitively makes sense: empirical observation reveals to any observer that natural language meaning is not fixed or absolute \citep{Harris1954, McDonaldShillcock2001} and interpretation itself is radically context-dependent \citep{Barsalou2003, YehBarsalou2006, Higgins1996}. Thus, any semantic meaning realized by an agent interpreting a natural language expression depends crucially on a combinatorially explosive set of potential factors. These factors include but are not limited to:
\begin{itemize}
    \item the surrounding sentential and discourse context \citep{BicknellEtAl2010, Kintsch2001}
    \item the agent's current attentional focus and task demands which can selectively highlight specific conceptual features, \citep{LeboisEtAl2015, YeeEtAl2012, HoenigEtAl2008, VanDantzigEtAl2008, BermeitingerEtAl2011, HsuEtAl2011}
    \item the agent's background knowledge, cultural milieu \citep{Athanasopoulos2009, ThompsonEtAl2020, Johns2021},
    \item transient psychological states \citep{CesarioEtAl2010} \footnote{While many consider LLMs simply statistical machines that are completely disconnected from the material reality of the physical world, it is certainly possible that real world physical variations in the environment that they inhabit or other random phenomena like cosmic rays could induce transient states. Additionally, when a model like Claude from Anthropic possesses knowledge of the time of the year, its having a French name and the training and system prompting of Claude reinforce a French mindset onto the model to the point that its response patterns and willingness to put in effort vary depending on whether the time of year coincides with French holidays.}
    \item the specific language being used, as different languages carve up semantic space differently \citep{MarianKaushanskaya2007}. 
\end{itemize}
This profound context dependence implies that meaning is not merely decoded from the specific words in and of themselves independently but that meaning is actively constructed or realized by an interpretive agent within a specific situation---\citep{Barsalou2009, PecherEtAl1998, BorghiEtAl2004}. The same expression presented to different agents---or to the same agent under different conditions---can yield different interpretations \citep{ConnellLynott2014, MuszThompsonSchill2015}. Critically, this highlights the \textit{observer-dependent} nature of the actualization of semantic meaning through direct interpretation, providing a quantum mechanical analog to an observable acting upon a state which we will explore the implications of more thoroughly in Section \ref{sec:quantum_semantics}. This kind of co-creative process has been posited to be mediated by a process called Relevance Realization (RR)—a core cognitive capacity enabling individuals to efficiently navigate vast semantic spaces by employing context-sensitive attentional mechanisms to identify relevant information and filter out the irrelevant \citep{VervaekeFerraro2013Relevance, andersen2022rr, jaeger_riedl_djedovic_vervaeke_walsh_2023}. 
Crucially, RR is non-algorithmic in the standard computational sense \citep{jaeger_riedl_djedovic_vervaeke_walsh_2023}, and as such is more adept at handling challenges like problem framing and indefinite search spaces that are intractable for purely formal systems operating in ``small worlds'' \citep{VervaekeRobinson2013Cognitive}. This embodied, observer-dependent, and contextually situated view of meaning construction challenges any notion of semantic expressions possessing intrinsic, context- or observer-independent meaning, aligning with dynamic models of semantic memory \citep{Barsalou2003, Barsalou2009, ConnellLynott2014, Hintzman1984, JamiesonEtAl2018, VanDamEtAl2010, LeboisEtAl2015, VanDamEtAl2012}.

The inherent uncertainties and deep context-dependencies of this dynamic process suggest that classical probabilistic and logical frameworks are insufficient. Consequently, researchers have turned to non-classical frameworks---such as those employing principles from quantum theory---to find more suitable mathematical tools. Such approaches have been adopted to model a wide range of cognitive phenomena, including concept combinations, decision-making, and memory \citep{Aerts2009, Bruza2009, Gabora2002, bruza2012, bruza2015, Yearsley2016, pothos2022}. The utility of these quantum-inspired models is not merely theoretical but is demonstrated in empirical studies. For instance, \citet{aerts2010pet} identified quantum-like contextuality effects when analyzing concept co-occurrence statistics for the ``Pet-Fish problem'' on the World-Wide Web, showing that meaning construction at this scale deviates from classical probabilistic assumptions. In a similar vein, other work has applied Bell's \citep{bell1964} theorem to human cognition, uncovering non-classical correlations that violate classical bounds in both information retrieval judgments \citep{uprety2020investigation} and cognitive decision-making tasks \citep{aerts2000, aerts2018_wind1, aerts2018_wind2}.

More recent work provides a crucial conceptual refinement by distinguishing between two types of contextual influence. In experiments on facial trait judgments, \citet{bruza2023} delineate between \textit{context-sensitivity}, a standard causal influence of context, and true \textit{contextuality}. The latter is defined as an acausal form of context dependence where a property may be genuinely indeterminate prior to measurement. They argue that if a cognitive phenomenon is found to be contextual, the underlying cognitive properties do not possess well-defined, pre-existing values. Instead, the property is actualized in the moment of judgment---a phenomenon that non-classical models are uniquely equipped to formalize.

Given the state of affairs, we aim in this work to accomplish two primary tasks: (1) identify the combinatorial problems that have stalled the apparent progress of frontier LLMs and (2) provide a practical path forward for understanding and studying natural language using a non-classical framework. To this end, in Section \ref{sec:kolmogorov_complexity}, we provide an information-theoretic exploration of the role of semantic degeneracy in single-turn problem solving tasks. Then, in Section \ref{sec:quantum_semantics}, we formulate a quantum semantic theoretical framework for the act of interpretation. Following this, we detail our experimental methodology in Section \ref{sec:experiments} to test whether natural language interpretation may exhibit non-classical behavior, leveraging LLMs as interpretative stand-ins. Section \ref{sec:results} presents our findings, and Section \ref{sec:discussion} discusses their broader implications for the future of computational linguistics and our understanding of cognitive science.

\section{Kolmogorov Complexity, Semantic Degeneracy, and the Challenge of Interpretation}
\label{sec:kolmogorov_complexity}

The concept of Kolmogorov complexity (KC) \citep{Kolmogorov65} provides a powerful lens through which to understand the fundamental limitations of natural language interpretation, particularly pertinent with respect to LLMs which are expected to reliably solve problems for users. Kolmogorov Complexity, $K(s)$, of a string $s$ is formally defined as the length of the shortest computer program (in a fixed universal description language) that produces $s$ as output. While KC strictly applies to finite strings, its underlying principle---the quantity of information required for minimal description---can be extended to the domain of semantics. For some semantic expression $S_{E}$ We can conceptualize $K(M(S_E))$ as the minimum number of bits required to unambiguously specify the \textit{intended meaning} $M(S_E)$ of a given semantic expression $S_E$. This specification must capture not only the identities of the concepts involved but also their precise contextual nuances and the intricate web of relationships binding them into a coherent whole. For instance, the KC of specifying a single, coherent interpretation of `I just went to the animal shelter and I brought a dog home' is low, whereas for a passage from a work as complicated and inter-connected as James Joyce's \textit{Finnegans Wake} e.g., 
\begin{quote}
And  what  sensitive  coin I'd  be  possessed  of  at  Latouche's,  begor,  I'd  sink  it  sumtotal,  every dolly  farting,  in  vestments  of  subdominal  poteen  at  prime  cost and  I  bait  you  my  chancey  oldcoat  against  the  whole  ounce  you half  on  your  backboard  (if  madamaud  strips  mesdamines  may cold  strafe  illglandsl)  that  I'm  the  gogetter  that'd  make  it  pay  like cash  registers  as  sure  as  there's  a  pot  on  a  pole.  And,  what  with  one man's  fish  and  a  dozen  men's  poissons,  sowing  my  wild  plums  to reap  ripe  plentihorns  mead,  lashings  of  erbole  and  hydromel  and bragget,  I'd  come  out  with  my  magic  fluke  in  close  time,  fair, free  and  frolicky,  zooming  tophole  on  the  mart  as  a  factor.
\end{quote}
 the KC would be extraordinarily high to the extent that the amount of constraints from context needed to disambiguate the `intended meaning' of the passage might require a program that is orders of magnitude longer than the original expression, reflecting the vast amount of information needed to resolve its multifaceted ambiguities into one particular reading.

Thus, the informational burden, $K(M(S_E))$, necessary to constrain interpretations to those closely aligned with an author's original intention, scales dramatically with the number and interconnectedness of the concepts and relationships within an expression $S_E$. While the constituent concepts $c_i$ form the foundation, each requires a number of bits for contextual instantiation. This linear relationship might be tractable on its own, but it is the relationships $r_{i,j}$ between concepts that contribute to actualizing the intended meaning, and specifying the interrelational constraints requires specifying an $O(N_{Concepts}^2)$ number of bits to have a shot at disambiguating potential interpretations and reproducing the intended one. Because of this, the KC for an expression can be written as an inequality with a lower bound based on the number 
\begin{equation} \label{eq:kc_components_detailed}
K(M(S_E)) \geq \sum_{i}^N c_{i} + \sum_{i,j}^{N}  \cdot c_{i,j}^2 \\
\end{equation}
As can be seen, $K(M(S_E))$ increases at least superlinearly as the complexity of an expression increases. Each of the $K(M(S_E))$ bits represents an informational specificity or a semantic choice point (e.g., deciding to interpret `bat' as an animal instead of a wooden stick used for baseball) that an interpreting agent must successfully reconstruct. Crucially, at each such decision point, there is a potential opportunity for degenerate solutions as many words have multiple meanings, and so we specify some degeneracy per bit $d_{b}$ (Note that this essentially serves as an error rate, but we employ it in this way to maintan the conceptual notion). Consequently, the probability of an interpreter correctly actualizing \textit{all} $K(M(S_E))$ bits to arrive at the precise intended meaning is the product of the probabilities of getting each bit right, 
\begin{equation}
P(\text{perfect interpretation}) = \frac{1}{N!} \prod_{k=1}^{K(M(S_E))} (\frac{1}{d})
\end{equation}

\begin{figure}[h!]
    \centering
    \includegraphics[width=\columnwidth]{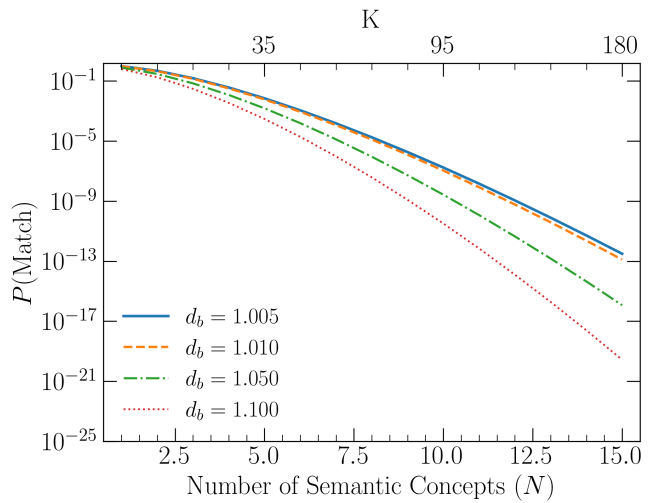} 
    \caption{Probability of perfect semantic interpretation versus the number of core semantic concepts ($N_{concepts}$) in an expression, illustrating the impact of superlinear growth in semantic complexity. The total semantic bits ($K$, shown on the top x-axis) are modeled as the sum of information required for individual concepts and their pairwise relationships, here assuming $c_{concept}=5$ bits per concept and $c_{relationship}=1$ bit per relationship. Different curves represent varying average probabilities of error per bit ($p_e$). The superlinear increase in $K$ with $N_{concepts}$ leads to a dramatically faster exponential decrease in the probability of flawless interpretation, underscoring the profound challenge posed by semantic degeneracy as inter-concept dependencies proliferate.}
    \label{fig:kc_vs_prob_interpretation}
\end{figure}

If we assume an average degeneracy per bit, $\bar{d_{b}}$
\begin{equation}
P(\text{perfect interpretation}) \approx \frac{1}{N!} ( \frac{1}{\bar{d_{b}}} )^{K(M(S_E))}
\end{equation}
a relationship which we illustrate graphically in 
Figure~\ref{fig:kc_vs_prob_interpretation}. 
As $K(M(S_E))$ grows in Figure~\ref{fig:kc_vs_prob_interpretation}, the probability of a perfect (or otherwise similar enough) interpretation diminishes exponentially, rapidly approaching zero for expressions of moderate complexity. This result provides a clear demonstration of semantic degeneracy in action: the combinatorial explosion of alternative, plausible interpretations surrounding any expression $S_{E}$. At this point, it is worth mentioning that this situation bears an analogy to concepts in statistical mechanics, wherein $S_E$ is like an ensemble of micro-states. An error in inferring even one bit (a constraint on a degree of freedom) leads to a different microstate, and, with the high dimensionality ($K(M(S_E))$), it becomes overwhelmingly probable that an interpreter will rarely, if ever, reproduce the specific set of micro-states 
(which we will argue in Section \ref{sec:quantum_semantics} are themselves unknowable a priori) that make up the ensemble, resulting in high `semantic entropy'. This KC-based analysis highlights a fundamental limitation for NLP systems and explains the persistent difficulties in LLM-assisted tasks requiring deep, unambiguous understanding or translation of semantically degenerate expressions: the LLM generates a \textit{plausible} meaning---one of many accessible microstates---but almost never the \textit{singularly intended} one. This result on its own highlights the need to move beyond training artificially intelligent systems that prioritize single-shot response success and to prioritize research on alternative models that can successfully simulate natural language in the way that LLMs have while also being able to dynamically update and adapt itself accordingly. It is our hope that these lessons here coupled with the quantum semantic approach can provide a clearer fundamental basis upon which future methods and models can be trained, tested, and evaluated.

\section{A theory of Quantum Semantics}
\label{sec:quantum_semantics}

The foundational premise of this work is that meaning is not an intrinsic, static property of a semantic expression, but rather an emergent phenomenon actualized through the dynamic interaction between the expression and an interpretive agent situated within a specific context. As described, this framework naturally challenges the assumption of realism which historically arose in classical physics. To formally model this observer-dependent and contextual nature of meaning, we propose a quantum semantic framework which mirrors the difference between quantum and classical physics. Semantic expressions are separated from syntax and instead treated as observables, mirroring how physical measurement outcomes in quantum systems are detached from the system as opposed to the realism as assumed in classical system. We hope that the processes of interpretation and meaning actualization are elucidated clearer under this quantum logical structure.

We begin by positing that a semantic expression, denoted $S_E$, does not possess a pre-defined, inherent meaning. Instead, it functions as a symbol that affords a spectrum of possible interpretations when engaged by an agent. To represent this capacity for interaction, $S_E$ is associated with a state vector $|\psi_{S_E}\rangle$ in a complex Hilbert space $\mathcal{H}_S$, the semantic state space. This vector is a linear superposition of basis states, $|\psi_{S_E}\rangle = \sum_i c_i |e_i\rangle$. A crucial aspect of this proposal is that the basis states $\{|e_i\rangle\}$ themselves are not assumed to be known or fixed a priori, nor are they necessarily interpretable as a universal set of predefined semantic primitives. Instead, they may represent abstract dimensions of potential semantic differentiation that only become operationally relevant or partially discernible through specific interpretive acts (measurements), where $\{|e_i\rangle\}$ is the set of eigenbasis of an observable $\hat{\mathcal{\mo}}$---here denoted by a Japanese character which is pronounced `mo'--- as posited by the agent performing the measurement, i.e.  interpreting the syntax. The Hilbert space $\mathcal{H}_S$ is thus a formal construct whose basis is defined by the set of all possible ways an expression can be distinctively engaged with, rather than by a pre-enumerated list of fundamental meanings. The coefficients $c_i$ are complex numbers ($c_i \in \mathbb{C}$), determined by how the semantic expression $S_E$ can be decomposed as a superposition of the semantic bases $\ket{e_i}$. While the squared moduli $|c_i|^2$ relate to probabilities of outcomes upon measurement with respect to a chosen observable, the complex phase of these coefficients holds additional information with no direct analogue in classical probabilistic models, and so it is impossible to derive any analogical particular meaning from these coefficients as one might consider weights in a language model. Indeed, the complex nature is fundamental for modeling potential interference and entanglement effects in semantic processing. The full interpretive significance of $c_i$ is only realized through the associated interpretive axis $\ket{e_i}$ set by an agent, i.e. there is no notion of pre-existing and isolated probabilities of meaning.

An agent $A$ engages $S_E$ through an interpretive observable, $\hat{\mathcal{\mo}}_A(t)$, which is dynamically constituted from its semantic memory (comprising goals, persona, knowledge, attentional state) as activated by the current context $C_A(t)$. The act of interpretation to ascertain a specific semantic aspect is represented by applying a Hermitian operator, denoted $\hat{\mathcal{\mo}}_A(t)$ to $|\psi_{S_E}\rangle$. This operator $\hat{\mathcal{\mo}}_A(t)$ embodies the specific semantic probe. The eigenvalues $\{m_i\}$ of $\hat{\mo}_A(t)$ encompass all possible outcomes of agent A's semantic measurement, specifically they represent all possible distinct interpretations actualized by the agent, even though they cannot be explicitly enumerated. The probability of actualizing interpretation $m_i$ is thus given by 
\begin{eqnarray}
P(m_i) = |\braket{c_i | \psi_{S_E}}|^2 = \bra{\psi_{S_E}}\hat{P}_i|\psi_{S_E}\rangle
\end{eqnarray}
where $\hat{P}_i=\ket{e_i}\bra{e_i}$ is the projection operator onto the eigenspace of $m_i$. This interaction is analogous to a quantum measurement.

With such formulation, semantic measurements by different agents can now possess non-commutativity, similar to quantum observables. If two distinct interpretive operations, $\hat{\mathcal{\mo}}_1$ and $\hat{\mathcal{\mo}}_2$, do not commute, i.e. $[\hat{\mathcal{\mo}}_1, \hat{\mathcal{\mo}}_2] \neq 0$, then the semantic aspects they probe cannot generally possess simultaneous, definite, pre-existing values.

The dynamics of meaning actualization extends to the time evolution of the interpretive process. The agent's interpretive observable $\hat{\mathcal{\mo}}(t)$ and the semantic expression $\ket{\psi_{S_E}(t)}$ are generally not static. While $\hat{\mathcal{\mo}}(t)$ as an observable can vary explicitly in time, $\ket{\psi_{S_E} (t)}$'s dynamics can, in analogy to quantum mechanics, be governed by a semantic Schrödinger equation:

\begin{equation}
i\hbar_{sem} \frac{\partial}{\partial t} \ket{\psi_{S_E}(t)} = \hat{H}_{sem}(t) \ket{\psi_{S_E}(t)}
\label{eq:schrodinger_semantic_condensed}
\end{equation}
where the semantic Hamiltonian, $\hat{H}_{sem}(t)$, generates this evolution, encapsulating drivers such as shifts in context $C_A(t)$, sequential information processing from $S_E$, and the agent's internal cognitive dynamics. The constant $\hbar_{sem}$ sets the scale for these semantic dynamics where quantum coherence between the bases ${\ket{e}_i}$ is important. This formalism allows for agents' interpretive engagement evolution. The dependence of $\hat{H}_{sem}(t)$ on $t$ allows for modeling changes in this evolution as the agent navigates different semantic loci. In this work, however, we set aside the time evolution component of our framework and focus more on certifying the quantum state nature of the semantic expression $S_E$ through sampling exploration by way of a Semantic Bell Test, thus adapting a logic that has previously been employed in cognitive psychological experiments to reveal non-classical contextuality in human judgements across diverse domains, including decision-making, information retrieval, and assessments of concepts \citep{aerts2000, aerts2018_wind1, aerts2018_wind2, uprety2020investigation}.

\section{Experimental Design}
\label{sec:experiments}

This section outlines the experimental methodology designed to test for non-classical correlations in semantic interpretation, analogous to a CHSH-type Bell test in quantum physics \citep{bell1964, clauser1969}. In particular, this experiment focuses on how Large Language Model (LLM) agents interpret ambiguous word pairs within simple sentence structures, under varying contextual (persona-based) settings.

\subsection{LLMs as observers}
In this work, LLM agents serve as the ``observers'' in this semantic Bell test. To mitigate potential model-specific biases and enhance the robustness of our findings, each agent instantiation is randomly selected from a predefined pool of diverse, state-of-the-art foundation models and providers. This pool includes models such as variants of Gemini (e.g., `gemini-1.5-flash', `gemini-2.0-flash-lite', `gemini-2.0-flash' ), Anthropic's Claude series (e.g., `claude-3-5-sonnet-latest', `claude-3-5-haiku-latest', and  `claude-3-7-sonnet-latest'), DeepSeek's `deepseek-chat', as well as various models from OpenAI (e.g., `gpt-4o`, `gpt-4o-mini', `gpt-4.1-mini', `gpt-4.1-nano', `gpt-4.1-nano').
This approach aligns with recommendations for multi-model triangulation to ensure fairness and generalizability of results \citep{GuoCaliskan2021, PayneEtAl2017, siddique2025, biasSurvey2023} and to ensure that any potential correlations we find are not restricted only to a single model's weights. For each experimental trial, two primary base personas, ``Alice'' and ``Bob'' are generated. These personas are characterized by randomly assigned attributes such as age (e.g., 25-70 years) and location (e.g., Bloomington, IN; Detroit, MI), which implicitly defines their primary language (English, in this setup). These attributes inform the agent's base semantic memory profile for the trial.

It should be noted here that, paradoxically, while LLMs exhibit limitations on complex tasks because of semantic degeneracy, their internal mechanisms—--specifically their attention architectures---function as a kind of black box which---like the brain---collapses the state of potential interpretations into a specific one that they use when responding to user inquiries.
Although the underlying mechanism is of course distinct from the biological and cognitive underpinnings of human linguistic interpretation, the similarity of the two `observing' a specific state suggests that LLMs do indeed effectively reproduce this function of language understanding and cognition. Thus, these models can serve as  experimental interpretative probes in natural language tasks. In addition, it has also been demonstrated that LLMs can generate responses that can to first order mimic human linguistic behavior in various contexts (e.g. example surveys \citep{kitadai2024,salencha2024, tjuatja2024}\footnote{The last of these references, \citet{tjuatja2024}, actually describes the performance of LLMs in this regard as poor as they note they are subject to perturbations that humans in a similar survey would not be as strongly affected. We note here that their findings here fit neatly within the framework of this work and that, if one additionally considers the role of relevance realization for the human subject (context-rich by way of physical embeddedness) versus the LLM (context-poor and disembodied), it appears clear that the main `problem' with the LLM responses in their experiment was due to a drastic contextual disadvantage.} \citet{kitadai2024} also note the power of personas in improving the verisimilitude of the responses, an important facet that we make use of as well. Thus, we argue, that one can and should use LLMs to probe the statistical patterns of semantic interpretations under diverse conditions (e.g. persona and context variations). By observing how LLMs grapple with semantic ambiguity and context-dependent meaning, we can gain insights into the mechanisms of interpretation and the types of computational strategies that are more or less successful in navigating tasks with high semantic degeneracy \citep{piantadosi2022meaning, lampinen-etal-2022-language}.

\subsection{Bell Test}
The core of our semantic Bell test involves presenting sentences containing two ambiguous words to LLM agents. The agents are then tasked with choosing a singular, unambiguous meanings for each of the words in the pair. Stimuli are constructed by embedding ambiguous word pairs (e.g., ``trunk'' with meanings `A' for storage/tree vs. ``bow'' with meanings `A' for ship front/knot) into simple sentence templates (e.g., ``The {word1} was settled near the {word2}''). Four distinct interpretive settings are defined for agents Alice (A, A') and Bob (B, B'), created by providing their base personas with additional, distinct short textual prompts designed to prime semantically orthogonal different contextual perspectives (e.g., ``You are a surgeon...'' vs. ``You are a bus driver...''). For each of these four settings, the corresponding LLM agent provides a singular simultaneous interpretation for both ambiguous words. Alice and Bob are not shown the definitions under consideration so as to avoid biasing them or limiting the semantic search space, and so a separate LLM call is required to then determine if each interpretation aligns with predefined meaning `$\alpha$' or `$\beta$', triggering a re-interpretation if the choice is unclear (e.g. if the two options for meaning for the word `chair' are `leader of a group' or `furniture to sit on' and the interpretation says `furniture or leader') or outside these options (e.g. for the chair example if it decides the sentence is referring to an execution by `electric chair', it would be considered outside of the definitional scope and a re-interpretation is carried out).\footnote{It is possible to imagine this experiment can be carried out with word choice sets with more than 2 meanings, and we plan to explore that in future work as this would allow us to further explore the qubit-style logic we might consider in more practical applications, but that is outside of the scope of this work.} These classified interpretations are mapped to numerical values (A $\rightarrow +1$, B $\rightarrow -1$), yielding a 2-element outcome vector for each setting. Finally, the outcome vectors ($\hat{\mathcal{\mo}_{A}}$, $\hat{\mathcal{\mo}_{A'}}$, $\hat{\mathcal{\mo}_{B}}$, $\hat{\mathcal{\mo}_{B'}}$) are normalized, and expectation values $E(XY)$ are calculated as the average dot product of corresponding pairs of ($\alpha$, $\beta$), which are then used to compute the CHSH S-value $S = E(A,B) - E(A,B') + E(A',B) + E(A',B')$. However, a critical assumption in this standard CHSH formulation is that of marginal consistency, also known as `no-signaling' \citep{dzhafarov2016contextuality, cervantes2017advanced}, which posits that the marginal probability for one agent's outcome is independent of the measurement setting of the other agent. As \citet{dzhafarov2016contextuality} have extensively discussed, this assumption is often violated in cognitive and behavioral experiments. Such violations, termed `inconsistent connectedness' can complicate the interpretation of Bell-type inequalities. Therefore, while a result of $|S| > 2$ suggests a violation of local realism, it must be interpreted with caution, as the excess correlation could potentially arise from these direct contextual influences rather than true contextuality.

All agent definitions and LLM response handling are carried out using the open-source \texttt{npcpy} package\footnote{\url{https://github.com/NPC-Worldwide/npcpy}}.
 \begin{table}
\begin{tabular}{c|c|c}
 Experiment & N Trials & |S| \\
     \hline
  1   & 5 & 2.8 \\     
  2   & 5 & 1.2 \\
  3   & 10 & 2.0 \\
  4   & 10 & 2.44 \\
  5   & 20 &  2.32 \\
  6   & 20 & 2.0 \\     
  7   & 50 & 2.33 \\
  8   & 200 & 1.83\\  
\end{tabular}
\caption{Calculated $|S|$ values for the CHSH inequality from different experimental runs, with slight variations between the persona configurations (age, location) as well as the total number of trials ($N$) per experiment. Values $|S|>2$ indicate a violation of the classical Bell-CHSH inequality.} 
  \label{tab:exp}
\end{table}

\section{Results}

\label{sec:results}

In this short section, we highlight and discuss the results of our Semantic Bell test. 

\begin{figure}[h!]
  \centering
  \includegraphics[width=\columnwidth]{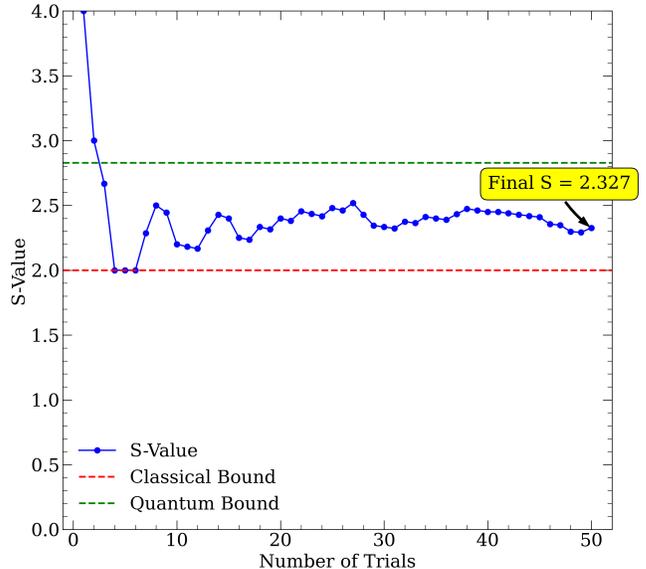}
\caption{Example evolution of the calculated CHSH S-value as the trial number progressed for an experiment run with a total of 50 trials, eventually settling at 2.327 (Experiment \#7 from Table~\ref{tab:exp}). The dashed red line at $S = \pm 2$ indicate the classical bounds of the CHSH inequality and the dashed green line indicates the proposed quantum bound at S=$2\sqrt2$.}  \label{fig:s_value_evolution_results}
 \end{figure}

In our experimentation, we conducted 8 experiments where we varied the number of trials in the individual experiments from 5-200 and instantiated different persona combinations, and we show the various $|S|$ values for the experiments in Table \ref{tab:exp}.  
For illustrative purposes, we show in Figure \ref{fig:s_value_evolution_results} the evolution of the |S| value's evolution across trial number for experiment \#7 with $N=50$ trials. To be clear, there is not any particular meaning to be derived from the variation in $|S|$ as it progresses with trial number as the $|S|$ value itself ought to be derived from an infinite number of trials. This illustration simply serves as a visual aid to show the stabilization of the $|S|$ value for a non-negligible number of trials.

Importantly in these results, we find a rich variety of correlation structure ranging from classical behavior to non-classical quantum logical. Theoretically, the upper bound for quantum systems in the Bell experiment is $|S|\leq2\sqrt{2}$. It is interesting then that one of our experiments (albeit $N=5$), the $|S|$ value reached 2.8, providing an empirical upper bound for these experiments that can be consistent with quantum computational frameworks.

\section{Discussion}
\label{sec:discussion}

The arguments presented in this work, concerning the fundamental limitations imposed by semantic degeneracy and the potential of a quantum semantic framework, offer a new lens through which to re-evaluate several core debates and prevailing methodologies in natural language processing and cognitive science. This discussion will first address how the information-theoretic challenge of semantic degeneracy, as quantified by Kolmogorov complexity, impacts the capabilities of distributional semantic models (DSMs) and Large Language Models (LLMs) in tasks requiring deep contextual understanding (Section \ref{ssec:degeneracy_dsm_llm}). Subsequently, we will explore how a quantum semantic framework, emphasizing observer-dependent meaning actualization, engages with and offers alternatives to classical theories of meaning and interpretation (Section \ref{ssec:qs_meaning_theories}). We will then connect these ideas to specific cognitive phenomena and psycholinguistic findings (Section \ref{ssec:qs_cognitive_connections}), and finally, consider the broader implications for NLP research (Section \ref{ssec:methodological_shifts}) and practical industrial applications (Section \ref{ssec:practical_architectures_hitl}).

\subsection{Semantic Degeneracy: Implications for DSMs and LLMs}
\label{ssec:degeneracy_dsm_llm}
Distributional Semantic Models, including contemporary Large Language Models (LLMs), have achieved remarkable success by learning statistical co-occurrences from vast text corpora \citep{Harris1954, mikolov2013distributed, DevlinEtAl2018, brown2020language}. These models implicitly operate under the assumption that sufficient statistical exposure can lead to robust representations of meaning. However, the principle of semantic degeneracy, particularly when analyzed through Kolmogorov complexity (KC) as detailed in Section \ref{sec:kolmogorov_complexity}, reveals a fundamental challenge to this assumption for tasks demanding precise, contextually-specific interpretations, as we have shown that as the complexity of a semantic expression and its required contextual disambiguation increases, the KC of specifying the singularly intended meaning grows superlinearly. This implies that any DSM or LLM---trained on finite sets of data and with fixed weights---will more likely than not provide solutions and interpretations that are not aligned with the `intended' one on some critical aspects that results in a complete breakdown in understanding. These information-theoretic limits stem from the combinatorial explosion of potential interpretations and can adequately explain the observed plateaus in LLM performance on complex reasoning tasks and the persistent difficulties DSMs face with ambiguous or context-rich texts \citep{Jones2018, HoffmanEtAl2013} despite increasing model size and data. Indeed it appears that the limitations of DSMs at navigating tasks with high ambiguity likely precludes them from ever achieving the status of `strong' AI. It appears likely that alternative methods or ensemble approaches (e.g. where LLMs are used as reasoning engines along with an alternative, more dynamic model) will prevail.

\subsection{Quantum Semantics versus Classical Theories of Meaning}
\label{ssec:qs_meaning_theories}
The limitations highlighted by semantic degeneracy motivate the exploration of alternative frameworks for meaning. Our proposed quantum semantic framework (Section \ref{sec:quantum_semantics}), which posits that meaning is not an intrinsic property of text but is actualized through an observer-dependent interpretive act, directly confronts the classical assumptions of semantic realism and locality often implicit in DSMs and traditional linguistic theories. Classical realism assumes pre-existing, definite meanings, while locality assumes semantic components can be independently determined. A quantum semantic approach, by contrast, treats expressions as affording a spectrum of potential interpretations (a superposition), with a specific meaning being `collapsed' or actualized by an agent's interpretive `measurement' within a given context. This aligns with constructivist theories in cognitive science that view understanding as an active, situated process \citep{Kintsch2001, Elman1990, Barsalou2003}. Philosophical critiques of essentialist views of meaning \citep[e.g., later][]{Wittgenstein1953, Quine1960} also resonate with our results as meaning becomes tied to use and interaction rather than inhering statically in symbols. The non-commutativity of interpretive operations---a feature of the quantum semantic model---implies that the order of contextual probing can affect the realized meaning, a phenomenon difficult to capture in purely classical additive models but observed in human cognition, e.g. order effects in judgment \citep{Hogarth1992} and quantum-like contextuality effects such as the `Pet-Fish' problem \citep{aerts2010pet}.

\subsection{Quantum Semantics: Connections to Cognitive and Psycholinguistic Phenomena}
\label{ssec:qs_cognitive_connections}
The observer-dependent actualization central to quantum semantics provides a compelling explanatory framework for a range of established cognitive and psycholinguistic findings. The profound influence of an agent's current context, goals, and primed semantic memory on interpretation \citep{AndersonMilson1989, Hintzman1984, Barsalou2009, JamiesonEtAl2018, ConnellLynott2014, MuszThompsonSchill2015} is naturally accommodated if meaning is not fixed but realized in interaction. This quantum concept of `measurement contextuality' aligns with models of dynamic semantic memory \citep{VanDamEtAl2010, LeboisEtAl2015, VanDamEtAl2012}, accounts of attentional focus \citep{YeeEtAl2012, HoenigEtAl2008, VanDantzigEtAl2008}, the non-algorithmic nature of Relevance Realization \citep{jaeger_riedl_djedovic_vervaeke_walsh_2023}, and even cross-linguistic variations in semantic categorization \citep{Athanasopoulos2009, MarianKaushanskaya2007, ThompsonEtAl2020}.

This notion of measurement contextuality finds strong empirical parallels in human judgment research. For instance, work by \citet{bruza2015, bruza2023} on facial judgments extends contextuality beyond language to perception, arguing that such properties are indeterminate and constructed in the moment of judgment. This reinforces our core premise: if concrete perceptual traits are indeterminate, then abstract semantic meaning is even more likely to require an interpretive act to become actualized. Our experimental findings provide direct quantitative support for this view. The violation of the CHSH inequality we observed is comparable in magnitude to results from cognitive experiments with humans, such as the work by \citet{aerts2018_wind1} on conceptual entanglement. This parallel suggests that LLMs tap into the same non-classical probabilistic structures inherent in human semantic processing.

This convergence of findings positions our LLM experiments as a novel methodology for investigating the fundamental nature of semantic contextuality itself. In this light, LLMs serve as sophisticated ``computational cognitive systems'' capable of bridging the gap between the macro-level statistical phenomena found in collective human data and the micro-level interpretive acts of an individual agent. The contextuality we observe, therefore, suggests this non-classical behavior is not a quirk of human psychology or a specific LLM architecture, but a pervasive feature of how semantic meaning is structured and processed in any complex, interconnected system.

\subsection{Methodological Shifts: Bayesian Exploration and Quantum Cognition Analogies}
\label{ssec:methodological_shifts}
The dual arguments---the limitations imposed by semantic degeneracy on classical systems and the potential of a quantum semantic framework---strongly suggest a paradigm shift in NLP methodologies towards non-classical, Bayesian-informed approaches. Instead of pursuing a single, definitive interpretation, techniques involving Monte Carlo sampling of interpretations under diverse contextual conditions, combined with dynamic explorations of semantic space (e.g., through Markovian random walks), may offer more practically useful and robust characterizations of text. This is particularly relevant for tasks such as nuanced translation, novel discovery, and complex single-turn completion where LLMs currently face inherent difficulties due to the Kolmogorov complexity challenge; a system designed to explore a wide distribution of plausible actualizations, effectively navigating many potential paths from problem to solution, while slower, would likely prove more effective at approximating an understanding of the requirements of the semantic expression. Adopting such a Bayesian-informed perspective allows systems to treat ambiguity not as an error to be eliminated, but as an inherent and informative feature of the semantic landscape. This approach directly addresses the computational intractability posed by semantic degeneracy, offering a practical path toward building more resilient and contextually-aware language technologies that better reflect the probabilistic nature of meaning itself.

\subsection{Practical Architectures and the Role of Human Oversight}
\label{ssec:practical_architectures_hitl}
Translating these conceptual shifts into practical applications, especially for complex scenarios such as multi-agent systems \citep{autogen2023, llamaindex2025}, or large-scale enterprise environments demanding dynamic document understanding \citep{sapAI2024, bcgERP2025}, necessitates the development of novel and adaptable system architectures for models attempting to approach intelligence. These architectures must be capable of managing and leveraging contextual variability rather than attempting to eliminate it. Crucially, the observer-dependent nature of meaning actualization, as posited by the quantum semantic framework, underscores the enduring and fundamental importance of Human-in-the-Loop (HITL) systems. Far from being a temporary measure until AI achieves perfect autonomy, HITL will remain an integral component for navigating inherent semantic ambiguity, validating interpretations, and ensuring that system outputs align with human goals and ethical considerations, particularly in safety-critical domains \citep{conformalLM2023, hmcf2025, labelyourdata2025} like healthcare, defense, and finance. The explicit recognition of both the fundamental limits of purely automated interpretation in open contexts and the potential of alternative, context-aware frameworks can guide the development of more realistic, robust, and ultimately more capable language technologies.

\section{Conclusions}
\label{sec:conclusions}

In this work, we have used a theoretical framework based on Kolmogorov complexity and a novel experimental design using Large Language Model agents to investigate the fundamental, non-classical nature of semantic meaning. We have approached this analysis by identifying the information-theoretic limits inherent in any act of linguistic interpretation and have provided the first known test of contextuality in the interpretive acts of diverse LLM-powered AI agents. Our major conclusions are the following:

\begin{enumerate}
    \item Semantic degeneracy is a fundamental property of natural language that imposes an information-theoretic limit on interpretation; our analysis using Kolmogorov complexity (Section \ref{sec:kolmogorov_complexity}) formalizes how this makes the recovery of a single intended meaning from a complex expression computationally intractable for any system, thereby providing a clear explanation for the observed performance plateaus in LLMs.
    
    \item Linguistic interpretation under ambiguity exhibits non-classical contextuality, as demonstrated by frequent and significant violations of the CHSH inequality ($|S|>2$) in our semantic Bell test experiments with LLM agents (Sections \ref{sec:experiments}, \ref{sec:results}).

    \item The contextuality measured in the interpretive acts of LLMs is consistent with a broader pattern of non-classical findings across human cognitive science, indicating that observer-dependence and indeterminacy are general principles of information processing and not simply artifacts of human psychology.
    
    \item The observer-dependent nature of meaning, confirmed by our experiment, reveals that there is no absolute, fundamental meaning to be found, only contextualized interpretations; consequently, the only viable scientific methodology is to shift from seeking any single ``correct'' answer and instead use repeated Bayesian-style sampling to characterize how these conditional interpretations interconnect within a possibility space.

    \item The consistent emergence of non-classical contextuality across a diverse pool of non-biological LLM agents, when considered alongside similar findings in human cognition, indicates that these statistical properties are not artifacts of any specific interpretive system but are objective, structural features of natural language itself.
\end{enumerate}



\bibliography{refs}

\begin{thebibliography}{89}%
\makeatletter
\providecommand \@ifxundefined [1]{%
 \@ifx{#1\undefined}
}%
\providecommand \@ifnum [1]{%
 \ifnum #1\expandafter \@firstoftwo
 \else \expandafter \@secondoftwo
 \fi
}%
\providecommand \@ifx [1]{%
 \ifx #1\expandafter \@firstoftwo
 \else \expandafter \@secondoftwo
 \fi
}%
\providecommand \natexlab [1]{#1}%
\providecommand \enquote  [1]{``#1''}%
\providecommand \bibnamefont  [1]{#1}%
\providecommand \bibfnamefont [1]{#1}%
\providecommand \citenamefont [1]{#1}%
\providecommand \href@noop [0]{\@secondoftwo}%
\providecommand \href [0]{\begingroup \@sanitize@url \@href}%
\providecommand \@href[1]{\@@startlink{#1}\@@href}%
\providecommand \@@href[1]{\endgroup#1\@@endlink}%
\providecommand \@sanitize@url [0]{\catcode `\\12\catcode `\$12\catcode `\&12\catcode `\#12\catcode `\^12\catcode `\_12\catcode `\%12\relax}%
\providecommand \@@startlink[1]{}%
\providecommand \@@endlink[0]{}%
\providecommand \url  [0]{\begingroup\@sanitize@url \@url }%
\providecommand \@url [1]{\endgroup\@href {#1}{\urlprefix }}%
\providecommand \urlprefix  [0]{URL }%
\providecommand \Eprint [0]{\href }%
\providecommand \doibase [0]{https://doi.org/}%
\providecommand \selectlanguage [0]{\@gobble}%
\providecommand \bibinfo  [0]{\@secondoftwo}%
\providecommand \bibfield  [0]{\@secondoftwo}%
\providecommand \translation [1]{[#1]}%
\providecommand \BibitemOpen [0]{}%
\providecommand \bibitemStop [0]{}%
\providecommand \bibitemNoStop [0]{.\EOS\space}%
\providecommand \EOS [0]{\spacefactor3000\relax}%
\providecommand \BibitemShut  [1]{\csname bibitem#1\endcsname}%
\let\auto@bib@innerbib\@empty
\bibitem [{\citenamefont {Dreyfus}(1992)}]{dreyfus1992}%
  \BibitemOpen
  \bibfield  {author} {\bibinfo {author} {\bibfnamefont {H.~L.}\ \bibnamefont {Dreyfus}},\ }\href@noop {} {\emph {\bibinfo {title} {What Computers Still Can?T Do: A Critique of Artificial Reason}}}\ (\bibinfo  {publisher} {MIT Press},\ \bibinfo {year} {1992})\BibitemShut {NoStop}%
\bibitem [{\citenamefont {Harris}(1954)}]{Harris1954}%
  \BibitemOpen
  \bibfield  {author} {\bibinfo {author} {\bibfnamefont {Z.~S.}\ \bibnamefont {Harris}},\ }\bibfield  {title} {\bibinfo {title} {{Distributional Structure}},\ }\href {https://doi.org/10.1080/00437956.1954.11659520} {\bibfield  {journal} {\bibinfo  {journal} {WORD}\ }\textbf {\bibinfo {volume} {10}},\ \bibinfo {pages} {146} (\bibinfo {year} {1954})}\BibitemShut {NoStop}%
\bibitem [{\citenamefont {Deerwester}\ \emph {et~al.}(1990)\citenamefont {Deerwester}, \citenamefont {Dumais}, \citenamefont {Furnas}, \citenamefont {Landauer},\ and\ \citenamefont {Harshman}}]{deerwester1990indexing}%
  \BibitemOpen
  \bibfield  {author} {\bibinfo {author} {\bibfnamefont {S.~C.}\ \bibnamefont {Deerwester}}, \bibinfo {author} {\bibfnamefont {S.~T.}\ \bibnamefont {Dumais}}, \bibinfo {author} {\bibfnamefont {G.~W.}\ \bibnamefont {Furnas}}, \bibinfo {author} {\bibfnamefont {T.~K.}\ \bibnamefont {Landauer}},\ and\ \bibinfo {author} {\bibfnamefont {R.~A.}\ \bibnamefont {Harshman}},\ }\bibfield  {title} {\bibinfo {title} {{Indexing by latent semantic analysis}},\ }\href@noop {} {\bibfield  {journal} {\bibinfo  {journal} {Journal of the American Society for Information Science}\ }\textbf {\bibinfo {volume} {41}},\ \bibinfo {pages} {391} (\bibinfo {year} {1990})}\BibitemShut {NoStop}%
\bibitem [{\citenamefont {Landauer}\ and\ \citenamefont {Dumais}(1997)}]{LandauerDumais1997}%
  \BibitemOpen
  \bibfield  {author} {\bibinfo {author} {\bibfnamefont {T.~K.}\ \bibnamefont {Landauer}}\ and\ \bibinfo {author} {\bibfnamefont {S.~T.}\ \bibnamefont {Dumais}},\ }\bibfield  {title} {\bibinfo {title} {{A Solution to Plato's Problem: The Latent Semantic Analysis Theory of Acquisition, Induction, and Representation of Knowledge}},\ }\href {https://doi.org/10.1037/0033-295X.104.2.211} {\bibfield  {journal} {\bibinfo  {journal} {Psychological Review}\ }\textbf {\bibinfo {volume} {104}},\ \bibinfo {pages} {211} (\bibinfo {year} {1997})}\BibitemShut {NoStop}%
\bibitem [{\citenamefont {Griffiths}\ \emph {et~al.}(2007)\citenamefont {Griffiths}, \citenamefont {Steyvers},\ and\ \citenamefont {Tenenbaum}}]{GriffithsEtAl2007}%
  \BibitemOpen
  \bibfield  {author} {\bibinfo {author} {\bibfnamefont {T.~L.}\ \bibnamefont {Griffiths}}, \bibinfo {author} {\bibfnamefont {M.}~\bibnamefont {Steyvers}},\ and\ \bibinfo {author} {\bibfnamefont {J.~B.}\ \bibnamefont {Tenenbaum}},\ }\bibfield  {title} {\bibinfo {title} {{Topics in semantic representation}},\ }\href {https://doi.org/10.1037/0033-295X.114.2.211} {\bibfield  {journal} {\bibinfo  {journal} {Psychological Review}\ }\textbf {\bibinfo {volume} {114}},\ \bibinfo {pages} {211} (\bibinfo {year} {2007})}\BibitemShut {NoStop}%
\bibitem [{\citenamefont {Mikolov}\ \emph {et~al.}(2013)\citenamefont {Mikolov}, \citenamefont {Sutskever}, \citenamefont {Chen}, \citenamefont {Corrado},\ and\ \citenamefont {Dean}}]{mikolov2013distributed}%
  \BibitemOpen
  \bibfield  {author} {\bibinfo {author} {\bibfnamefont {T.}~\bibnamefont {Mikolov}}, \bibinfo {author} {\bibfnamefont {I.}~\bibnamefont {Sutskever}}, \bibinfo {author} {\bibfnamefont {K.}~\bibnamefont {Chen}}, \bibinfo {author} {\bibfnamefont {G.~S.}\ \bibnamefont {Corrado}},\ and\ \bibinfo {author} {\bibfnamefont {J.}~\bibnamefont {Dean}},\ }\href@noop {} {\bibinfo {title} {{Distributed Representations of Words and Phrases and their Compositionality}}},\ \bibinfo {howpublished} {arXiv preprint arXiv:1310.4546} (\bibinfo {year} {2013}),\ \Eprint {https://arxiv.org/abs/1310.4546} {arXiv:1310.4546 [cs.CL]} \BibitemShut {NoStop}%
\bibitem [{\citenamefont {Devlin}\ \emph {et~al.}(2018)\citenamefont {Devlin}, \citenamefont {Chang}, \citenamefont {Lee},\ and\ \citenamefont {Toutanova}}]{DevlinEtAl2018}%
  \BibitemOpen
  \bibfield  {author} {\bibinfo {author} {\bibfnamefont {J.}~\bibnamefont {Devlin}}, \bibinfo {author} {\bibfnamefont {M.-W.}\ \bibnamefont {Chang}}, \bibinfo {author} {\bibfnamefont {K.}~\bibnamefont {Lee}},\ and\ \bibinfo {author} {\bibfnamefont {K.}~\bibnamefont {Toutanova}},\ }\href@noop {} {\bibinfo {title} {{Bert: Pre-training of deep bidirectional transformers for language understanding}}},\ \bibinfo {howpublished} {arXiv Preprint arXiv:1810.04805} (\bibinfo {year} {2018}),\ \Eprint {https://arxiv.org/abs/1810.04805} {arXiv:1810.04805} \BibitemShut {NoStop}%
\bibitem [{\citenamefont {Radford}\ \emph {et~al.}(2019)\citenamefont {Radford}, \citenamefont {Wu}, \citenamefont {Child}, \citenamefont {Luan}, \citenamefont {Amodei},\ and\ \citenamefont {Sutskever}}]{RadfordEtAl2019gpt2}%
  \BibitemOpen
  \bibfield  {author} {\bibinfo {author} {\bibfnamefont {A.}~\bibnamefont {Radford}}, \bibinfo {author} {\bibfnamefont {J.}~\bibnamefont {Wu}}, \bibinfo {author} {\bibfnamefont {R.}~\bibnamefont {Child}}, \bibinfo {author} {\bibfnamefont {D.}~\bibnamefont {Luan}}, \bibinfo {author} {\bibfnamefont {D.}~\bibnamefont {Amodei}},\ and\ \bibinfo {author} {\bibfnamefont {I.}~\bibnamefont {Sutskever}},\ }\href@noop {} {\bibinfo {title} {{Language models are unsupervised multitask learners}}},\ \bibinfo {howpublished} {OpenAI Blog} (\bibinfo {year} {2019}),\ \bibinfo {note} {openAI technical report, Version 1}\BibitemShut {NoStop}%
\bibitem [{\citenamefont {Brown}\ \emph {et~al.}(2020)\citenamefont {Brown}, \citenamefont {Mann}, \citenamefont {Ryder}, \citenamefont {Subbiah}, \citenamefont {Kaplan}, \citenamefont {Dhariwal}, \citenamefont {Neelakantan}, \citenamefont {Shyam}, \citenamefont {Sastry}, \citenamefont {Askell} \emph {et~al.}}]{brown2020language}%
  \BibitemOpen
  \bibfield  {author} {\bibinfo {author} {\bibfnamefont {T.}~\bibnamefont {Brown}}, \bibinfo {author} {\bibfnamefont {B.}~\bibnamefont {Mann}}, \bibinfo {author} {\bibfnamefont {N.}~\bibnamefont {Ryder}}, \bibinfo {author} {\bibfnamefont {M.}~\bibnamefont {Subbiah}}, \bibinfo {author} {\bibfnamefont {J.~D.}\ \bibnamefont {Kaplan}}, \bibinfo {author} {\bibfnamefont {P.}~\bibnamefont {Dhariwal}}, \bibinfo {author} {\bibfnamefont {A.}~\bibnamefont {Neelakantan}}, \bibinfo {author} {\bibfnamefont {P.}~\bibnamefont {Shyam}}, \bibinfo {author} {\bibfnamefont {G.}~\bibnamefont {Sastry}}, \bibinfo {author} {\bibfnamefont {A.}~\bibnamefont {Askell}}, \emph {et~al.},\ }\bibfield  {title} {\bibinfo {title} {{Language models are few-shot learners}},\ }in\ \href@noop {} {\emph {\bibinfo {booktitle} {Advances in Neural Information Processing Systems 33 (NeurIPS 2020)}}},\ \bibinfo {editor} {edited by\ \bibinfo {editor} {\bibfnamefont {H.}~\bibnamefont {Larochelle}}, \bibinfo {editor} {\bibfnamefont {M.}~\bibnamefont
  {Ranzato}}, \bibinfo {editor} {\bibfnamefont {R.}~\bibnamefont {Hadsell}}, \bibinfo {editor} {\bibfnamefont {M.~F.}\ \bibnamefont {Balcan}},\ and\ \bibinfo {editor} {\bibfnamefont {H.}~\bibnamefont {Lin}}}\ (\bibinfo  {publisher} {Curran Associates, Inc.},\ \bibinfo {year} {2020})\ pp.\ \bibinfo {pages} {1877--1901}\BibitemShut {NoStop}%
\bibitem [{\citenamefont {Vaswani}\ \emph {et~al.}(2017)\citenamefont {Vaswani}, \citenamefont {Shazeer}, \citenamefont {Parmar}, \citenamefont {Uszkoreit}, \citenamefont {Jones}, \citenamefont {Gomez}, \citenamefont {Kaiser},\ and\ \citenamefont {Polosukhin}}]{vaswani2017attention}%
  \BibitemOpen
  \bibfield  {author} {\bibinfo {author} {\bibfnamefont {A.}~\bibnamefont {Vaswani}}, \bibinfo {author} {\bibfnamefont {N.}~\bibnamefont {Shazeer}}, \bibinfo {author} {\bibfnamefont {N.}~\bibnamefont {Parmar}}, \bibinfo {author} {\bibfnamefont {J.}~\bibnamefont {Uszkoreit}}, \bibinfo {author} {\bibfnamefont {L.}~\bibnamefont {Jones}}, \bibinfo {author} {\bibfnamefont {A.~N.}\ \bibnamefont {Gomez}}, \bibinfo {author} {\bibfnamefont {L.}~\bibnamefont {Kaiser}},\ and\ \bibinfo {author} {\bibfnamefont {I.}~\bibnamefont {Polosukhin}},\ }\href@noop {} {\bibinfo {title} {{Attention Is All You Need}}},\ \bibinfo {howpublished} {arXiv preprint arXiv:1706.03762} (\bibinfo {year} {2017}),\ \Eprint {https://arxiv.org/abs/1706.03762} {arXiv:1706.03762 [cs.CL]} \BibitemShut {NoStop}%
\bibitem [{\citenamefont {Jones}(2018)}]{Jones2018}%
  \BibitemOpen
  \bibfield  {author} {\bibinfo {author} {\bibfnamefont {M.~N.}\ \bibnamefont {Jones}},\ }\bibfield  {title} {\bibinfo {title} {{When does abstraction occur in semantic memory: Insights from distributional models}},\ }\href@noop {} {\bibfield  {journal} {\bibinfo  {journal} {Language, Cognition and Neuroscience}\ }\textbf {\bibinfo {volume} {34}},\ \bibinfo {pages} {1338} (\bibinfo {year} {2018})}\BibitemShut {NoStop}%
\bibitem [{\citenamefont {Hoffman}\ \emph {et~al.}(2013)\citenamefont {Hoffman}, \citenamefont {Lambon~Ralph},\ and\ \citenamefont {Rogers}}]{HoffmanEtAl2013}%
  \BibitemOpen
  \bibfield  {author} {\bibinfo {author} {\bibfnamefont {P.}~\bibnamefont {Hoffman}}, \bibinfo {author} {\bibfnamefont {M.~A.}\ \bibnamefont {Lambon~Ralph}},\ and\ \bibinfo {author} {\bibfnamefont {T.~T.}\ \bibnamefont {Rogers}},\ }\bibfield  {title} {\bibinfo {title} {{Semantic diversity: A measure of semantic ambiguity based on variability in the contextual usage of words}},\ }\href {https://doi.org/10.3758/s13428-012-0278-x} {\bibfield  {journal} {\bibinfo  {journal} {Behavior Research Methods}\ }\textbf {\bibinfo {volume} {45}},\ \bibinfo {pages} {718} (\bibinfo {year} {2013})}\BibitemShut {NoStop}%
\bibitem [{\citenamefont {Assran}\ \emph {et~al.}(2023)\citenamefont {Assran}, \citenamefont {Duval}, \citenamefont {Misra}, \citenamefont {Bojanowski}, \citenamefont {Vincent}, \citenamefont {Rabbat}, \citenamefont {LeCun},\ and\ \citenamefont {Ballas}}]{Assran_2023_CVPR}%
  \BibitemOpen
  \bibfield  {author} {\bibinfo {author} {\bibfnamefont {M.}~\bibnamefont {Assran}}, \bibinfo {author} {\bibfnamefont {Q.}~\bibnamefont {Duval}}, \bibinfo {author} {\bibfnamefont {I.}~\bibnamefont {Misra}}, \bibinfo {author} {\bibfnamefont {P.}~\bibnamefont {Bojanowski}}, \bibinfo {author} {\bibfnamefont {P.}~\bibnamefont {Vincent}}, \bibinfo {author} {\bibfnamefont {M.}~\bibnamefont {Rabbat}}, \bibinfo {author} {\bibfnamefont {Y.}~\bibnamefont {LeCun}},\ and\ \bibinfo {author} {\bibfnamefont {N.}~\bibnamefont {Ballas}},\ }\bibfield  {title} {\bibinfo {title} {Self-supervised learning from images with a joint-embedding predictive architecture},\ }in\ \href@noop {} {\emph {\bibinfo {booktitle} {Proceedings of the IEEE/CVF Conference on Computer Vision and Pattern Recognition (CVPR)}}}\ (\bibinfo {year} {2023})\ pp.\ \bibinfo {pages} {15619--15629}\BibitemShut {NoStop}%
\bibitem [{\citenamefont {{Gu}}\ and\ \citenamefont {{Dao}}(2023)}]{gu_mamba_2023}%
  \BibitemOpen
  \bibfield  {author} {\bibinfo {author} {\bibfnamefont {A.}~\bibnamefont {{Gu}}}\ and\ \bibinfo {author} {\bibfnamefont {T.}~\bibnamefont {{Dao}}},\ }\bibfield  {title} {\bibinfo {title} {{Mamba: Linear-Time Sequence Modeling with Selective State Spaces}},\ }\href {https://doi.org/10.48550/arXiv.2312.00752} {\bibfield  {journal} {\bibinfo  {journal} {arXiv e-prints}\ ,\ \bibinfo {eid} {arXiv:2312.00752}} (\bibinfo {year} {2023})},\ \Eprint {https://arxiv.org/abs/2312.00752} {arXiv:2312.00752 [cs.LG]} \BibitemShut {NoStop}%
\bibitem [{\citenamefont {{Darlow}}\ \emph {et~al.}(2025)\citenamefont {{Darlow}}, \citenamefont {{Regan}}, \citenamefont {{Risi}}, \citenamefont {{Seely}},\ and\ \citenamefont {{Jones}}}]{darlow_ctm_2025}%
  \BibitemOpen
  \bibfield  {author} {\bibinfo {author} {\bibfnamefont {L.}~\bibnamefont {{Darlow}}}, \bibinfo {author} {\bibfnamefont {C.}~\bibnamefont {{Regan}}}, \bibinfo {author} {\bibfnamefont {S.}~\bibnamefont {{Risi}}}, \bibinfo {author} {\bibfnamefont {J.}~\bibnamefont {{Seely}}},\ and\ \bibinfo {author} {\bibfnamefont {L.}~\bibnamefont {{Jones}}},\ }\bibfield  {title} {\bibinfo {title} {{Continuous Thought Machines}},\ }\href {https://doi.org/10.48550/arXiv.2505.05522} {\bibfield  {journal} {\bibinfo  {journal} {arXiv e-prints}\ ,\ \bibinfo {eid} {arXiv:2505.05522}} (\bibinfo {year} {2025})},\ \Eprint {https://arxiv.org/abs/2505.05522} {arXiv:2505.05522 [stat.ML]} \BibitemShut {NoStop}%
\bibitem [{\citenamefont {{Sun}}\ \emph {et~al.}(2025)\citenamefont {{Sun}}, \citenamefont {{Cetin}},\ and\ \citenamefont {{Tang}}}]{transformer2}%
  \BibitemOpen
  \bibfield  {author} {\bibinfo {author} {\bibfnamefont {Q.}~\bibnamefont {{Sun}}}, \bibinfo {author} {\bibfnamefont {E.}~\bibnamefont {{Cetin}}},\ and\ \bibinfo {author} {\bibfnamefont {Y.}~\bibnamefont {{Tang}}},\ }\bibfield  {title} {\bibinfo {title} {{Transformer-Squared: Self-adaptive LLMs}},\ }\href {https://doi.org/10.48550/arXiv.2501.06252} {\bibfield  {journal} {\bibinfo  {journal} {arXiv e-prints}\ ,\ \bibinfo {eid} {arXiv:2501.06252}} (\bibinfo {year} {2025})},\ \Eprint {https://arxiv.org/abs/2501.06252} {arXiv:2501.06252 [cs.LG]} \BibitemShut {NoStop}%
\bibitem [{\citenamefont {{Simonds}}\ and\ \citenamefont {{Yoshiyama}}(2025)}]{simonds2025}%
  \BibitemOpen
  \bibfield  {author} {\bibinfo {author} {\bibfnamefont {T.}~\bibnamefont {{Simonds}}}\ and\ \bibinfo {author} {\bibfnamefont {A.}~\bibnamefont {{Yoshiyama}}},\ }\bibfield  {title} {\bibinfo {title} {{LADDER: Self-Improving LLMs Through Recursive Problem Decomposition}},\ }\href {https://doi.org/10.48550/arXiv.2503.00735} {\bibfield  {journal} {\bibinfo  {journal} {arXiv e-prints}\ ,\ \bibinfo {eid} {arXiv:2503.00735}} (\bibinfo {year} {2025})},\ \Eprint {https://arxiv.org/abs/2503.00735} {arXiv:2503.00735 [cs.LG]} \BibitemShut {NoStop}%
\bibitem [{\citenamefont {Tabossi}\ \emph {et~al.}(1987)\citenamefont {Tabossi}, \citenamefont {Colombo},\ and\ \citenamefont {Job}}]{TabossiEtAl1987}%
  \BibitemOpen
  \bibfield  {author} {\bibinfo {author} {\bibfnamefont {P.}~\bibnamefont {Tabossi}}, \bibinfo {author} {\bibfnamefont {L.}~\bibnamefont {Colombo}},\ and\ \bibinfo {author} {\bibfnamefont {R.}~\bibnamefont {Job}},\ }\bibfield  {title} {\bibinfo {title} {{Accessing lexical ambiguity: Effects of context and dominance}},\ }\href {https://doi.org/10.1007/BF00308682} {\bibfield  {journal} {\bibinfo  {journal} {Psychological Research}\ }\textbf {\bibinfo {volume} {49}},\ \bibinfo {pages} {161} (\bibinfo {year} {1987})}\BibitemShut {NoStop}%
\bibitem [{\citenamefont {Binder}\ and\ \citenamefont {Rayner}(1998)}]{BinderRayner1998}%
  \BibitemOpen
  \bibfield  {author} {\bibinfo {author} {\bibfnamefont {K.~S.}\ \bibnamefont {Binder}}\ and\ \bibinfo {author} {\bibfnamefont {K.}~\bibnamefont {Rayner}},\ }\bibfield  {title} {\bibinfo {title} {{Contextual strength does not modulate the subordinate bias effect: Evidence from eye fixations and self-paced reading}},\ }\href {https://doi.org/10.3758/BF03212950} {\bibfield  {journal} {\bibinfo  {journal} {Psychonomic Bulletin \& Review}\ }\textbf {\bibinfo {volume} {5}},\ \bibinfo {pages} {271} (\bibinfo {year} {1998})}\BibitemShut {NoStop}%
\bibitem [{\citenamefont {Rayner}\ \emph {et~al.}(2006)\citenamefont {Rayner}, \citenamefont {Cook}, \citenamefont {Juhasz},\ and\ \citenamefont {Frazier}}]{RaynerEtAl2006}%
  \BibitemOpen
  \bibfield  {author} {\bibinfo {author} {\bibfnamefont {K.}~\bibnamefont {Rayner}}, \bibinfo {author} {\bibfnamefont {A.~E.}\ \bibnamefont {Cook}}, \bibinfo {author} {\bibfnamefont {B.~J.}\ \bibnamefont {Juhasz}},\ and\ \bibinfo {author} {\bibfnamefont {L.}~\bibnamefont {Frazier}},\ }\bibfield  {title} {\bibinfo {title} {{Immediate disambiguation of lexically ambiguous words during reading: Evidence from eye movements}},\ }\href {https://doi.org/10.1348/000712605X89363} {\bibfield  {journal} {\bibinfo  {journal} {British Journal of Psychology}\ }\textbf {\bibinfo {volume} {97}},\ \bibinfo {pages} {467} (\bibinfo {year} {2006})}\BibitemShut {NoStop}%
\bibitem [{\citenamefont {McDonald}\ and\ \citenamefont {Shillcock}(2001)}]{McDonaldShillcock2001}%
  \BibitemOpen
  \bibfield  {author} {\bibinfo {author} {\bibfnamefont {S.~A.}\ \bibnamefont {McDonald}}\ and\ \bibinfo {author} {\bibfnamefont {R.~C.}\ \bibnamefont {Shillcock}},\ }\bibfield  {title} {\bibinfo {title} {{Rethinking the Word Frequency Effect: The Neglected Role of Distributional Information in Lexical Processing}},\ }\href {https://doi.org/10.1177/00238309010440030101} {\bibfield  {journal} {\bibinfo  {journal} {Language and Speech}\ }\textbf {\bibinfo {volume} {44}},\ \bibinfo {pages} {295} (\bibinfo {year} {2001})}\BibitemShut {NoStop}%
\bibitem [{\citenamefont {Barsalou}(2003)}]{Barsalou2003}%
  \BibitemOpen
  \bibfield  {author} {\bibinfo {author} {\bibfnamefont {L.~W.}\ \bibnamefont {Barsalou}},\ }\bibfield  {title} {\bibinfo {title} {{Situated simulation in the human conceptual system}},\ }\href {https://doi.org/10.1080/01690960344000026} {\bibfield  {journal} {\bibinfo  {journal} {Language and Cognitive Processes}\ }\textbf {\bibinfo {volume} {18}},\ \bibinfo {pages} {513} (\bibinfo {year} {2003})}\BibitemShut {NoStop}%
\bibitem [{\citenamefont {Yeh}\ and\ \citenamefont {Barsalou}(2006)}]{YehBarsalou2006}%
  \BibitemOpen
  \bibfield  {author} {\bibinfo {author} {\bibfnamefont {W.}~\bibnamefont {Yeh}}\ and\ \bibinfo {author} {\bibfnamefont {L.~W.}\ \bibnamefont {Barsalou}},\ }\bibfield  {title} {\bibinfo {title} {{The Situated Nature of Concepts}},\ }\href {https://doi.org/10.2307/20445349} {\bibfield  {journal} {\bibinfo  {journal} {The American Journal of Psychology}\ }\textbf {\bibinfo {volume} {119}},\ \bibinfo {pages} {349} (\bibinfo {year} {2006})}\BibitemShut {NoStop}%
\bibitem [{\citenamefont {Higgins}(1996)}]{Higgins1996}%
  \BibitemOpen
  \bibfield  {author} {\bibinfo {author} {\bibfnamefont {E.~T.}\ \bibnamefont {Higgins}},\ }\bibfield  {title} {\bibinfo {title} {{Knowledge activation: Accessibility, applicability, and salience}},\ }in\ \href@noop {} {\emph {\bibinfo {booktitle} {Social Psychology: Handbook of Basic Principles}}},\ \bibinfo {editor} {edited by\ \bibinfo {editor} {\bibfnamefont {E.~T.}\ \bibnamefont {Higgins}}\ and\ \bibinfo {editor} {\bibfnamefont {A.~W.}\ \bibnamefont {Kruglanski}}}\ (\bibinfo  {publisher} {Guilford Press},\ \bibinfo {year} {1996})\ pp.\ \bibinfo {pages} {133--168}\BibitemShut {NoStop}%
\bibitem [{\citenamefont {Bicknell}\ \emph {et~al.}(2010)\citenamefont {Bicknell}, \citenamefont {Elman}, \citenamefont {Hare}, \citenamefont {McRae},\ and\ \citenamefont {Kutas}}]{BicknellEtAl2010}%
  \BibitemOpen
  \bibfield  {author} {\bibinfo {author} {\bibfnamefont {K.}~\bibnamefont {Bicknell}}, \bibinfo {author} {\bibfnamefont {J.~L.}\ \bibnamefont {Elman}}, \bibinfo {author} {\bibfnamefont {M.}~\bibnamefont {Hare}}, \bibinfo {author} {\bibfnamefont {K.}~\bibnamefont {McRae}},\ and\ \bibinfo {author} {\bibfnamefont {M.}~\bibnamefont {Kutas}},\ }\bibfield  {title} {\bibinfo {title} {{Effects of event knowledge in processing verbal arguments}},\ }\href {https://doi.org/10.1016/j.jml.2010.08.004} {\bibfield  {journal} {\bibinfo  {journal} {Journal of Memory and Language}\ }\textbf {\bibinfo {volume} {63}},\ \bibinfo {pages} {489} (\bibinfo {year} {2010})}\BibitemShut {NoStop}%
\bibitem [{\citenamefont {Kintsch}(2001)}]{Kintsch2001}%
  \BibitemOpen
  \bibfield  {author} {\bibinfo {author} {\bibfnamefont {W.}~\bibnamefont {Kintsch}},\ }\bibfield  {title} {\bibinfo {title} {{Predication}},\ }\href {https://doi.org/10.1207/s15516709cog2502_1} {\bibfield  {journal} {\bibinfo  {journal} {Cognitive Science}\ }\textbf {\bibinfo {volume} {25}},\ \bibinfo {pages} {173} (\bibinfo {year} {2001})}\BibitemShut {NoStop}%
\bibitem [{\citenamefont {Lebois}\ \emph {et~al.}(2015)\citenamefont {Lebois}, \citenamefont {Wilson-Mendenhall},\ and\ \citenamefont {Barsalou}}]{LeboisEtAl2015}%
  \BibitemOpen
  \bibfield  {author} {\bibinfo {author} {\bibfnamefont {L.~A.~M.}\ \bibnamefont {Lebois}}, \bibinfo {author} {\bibfnamefont {C.~D.}\ \bibnamefont {Wilson-Mendenhall}},\ and\ \bibinfo {author} {\bibfnamefont {L.~W.}\ \bibnamefont {Barsalou}},\ }\bibfield  {title} {\bibinfo {title} {{Are Automatic Conceptual Cores the Gold Standard of Semantic Processing? The Context-Dependence of Spatial Meaning in Grounded Congruency Effects}},\ }\href {https://doi.org/10.1111/cogs.12174} {\bibfield  {journal} {\bibinfo  {journal} {Cognitive Science}\ }\textbf {\bibinfo {volume} {39}},\ \bibinfo {pages} {1764} (\bibinfo {year} {2015})}\BibitemShut {NoStop}%
\bibitem [{\citenamefont {Yee}\ \emph {et~al.}(2012)\citenamefont {Yee}, \citenamefont {Ahmed},\ and\ \citenamefont {Thompson-Schill}}]{YeeEtAl2012}%
  \BibitemOpen
  \bibfield  {author} {\bibinfo {author} {\bibfnamefont {E.}~\bibnamefont {Yee}}, \bibinfo {author} {\bibfnamefont {S.~Z.}\ \bibnamefont {Ahmed}},\ and\ \bibinfo {author} {\bibfnamefont {S.~L.}\ \bibnamefont {Thompson-Schill}},\ }\bibfield  {title} {\bibinfo {title} {{Colorless green ideas (can) prime furiously}},\ }\href@noop {} {\bibfield  {journal} {\bibinfo  {journal} {Psychological Science}\ }\textbf {\bibinfo {volume} {23}},\ \bibinfo {pages} {364} (\bibinfo {year} {2012})}\BibitemShut {NoStop}%
\bibitem [{\citenamefont {Hoenig}\ \emph {et~al.}(2008)\citenamefont {Hoenig}, \citenamefont {Sim}, \citenamefont {Bochev}, \citenamefont {Herrnberger},\ and\ \citenamefont {Kiefer}}]{HoenigEtAl2008}%
  \BibitemOpen
  \bibfield  {author} {\bibinfo {author} {\bibfnamefont {K.}~\bibnamefont {Hoenig}}, \bibinfo {author} {\bibfnamefont {E.-J.}\ \bibnamefont {Sim}}, \bibinfo {author} {\bibfnamefont {V.}~\bibnamefont {Bochev}}, \bibinfo {author} {\bibfnamefont {B.}~\bibnamefont {Herrnberger}},\ and\ \bibinfo {author} {\bibfnamefont {M.}~\bibnamefont {Kiefer}},\ }\bibfield  {title} {\bibinfo {title} {{Conceptual Flexibility in the Human Brain: Dynamic Recruitment of Semantic Maps from Visual, Motor, and Motion-related Areas}},\ }\href {https://doi.org/10.1162/jocn.2008.20123} {\bibfield  {journal} {\bibinfo  {journal} {Journal of Cognitive Neuroscience}\ }\textbf {\bibinfo {volume} {20}},\ \bibinfo {pages} {1799} (\bibinfo {year} {2008})}\BibitemShut {NoStop}%
\bibitem [{\citenamefont {Van~Dantzig}\ \emph {et~al.}(2008)\citenamefont {Van~Dantzig}, \citenamefont {Pecher}, \citenamefont {Zeelenberg},\ and\ \citenamefont {Barsalou}}]{VanDantzigEtAl2008}%
  \BibitemOpen
  \bibfield  {author} {\bibinfo {author} {\bibfnamefont {S.}~\bibnamefont {Van~Dantzig}}, \bibinfo {author} {\bibfnamefont {D.}~\bibnamefont {Pecher}}, \bibinfo {author} {\bibfnamefont {R.}~\bibnamefont {Zeelenberg}},\ and\ \bibinfo {author} {\bibfnamefont {L.~W.}\ \bibnamefont {Barsalou}},\ }\bibfield  {title} {\bibinfo {title} {{Perceptual Processing Affects Conceptual Processing}},\ }\href {https://doi.org/10.1080/03640210802035365} {\bibfield  {journal} {\bibinfo  {journal} {Cognitive Science}\ }\textbf {\bibinfo {volume} {32}},\ \bibinfo {pages} {579} (\bibinfo {year} {2008})}\BibitemShut {NoStop}%
\bibitem [{\citenamefont {Bermeitinger}\ \emph {et~al.}(2011)\citenamefont {Bermeitinger}, \citenamefont {Wentura},\ and\ \citenamefont {Frings}}]{BermeitingerEtAl2011}%
  \BibitemOpen
  \bibfield  {author} {\bibinfo {author} {\bibfnamefont {C.}~\bibnamefont {Bermeitinger}}, \bibinfo {author} {\bibfnamefont {D.}~\bibnamefont {Wentura}},\ and\ \bibinfo {author} {\bibfnamefont {C.}~\bibnamefont {Frings}},\ }\bibfield  {title} {\bibinfo {title} {{How to switch on and switch off semantic priming effects: Activation processes in category memory depend on focusing specific feature dimensions}},\ }\href@noop {} {\bibfield  {journal} {\bibinfo  {journal} {Psychonomic Bulletin \& Review}\ }\textbf {\bibinfo {volume} {18}},\ \bibinfo {pages} {579} (\bibinfo {year} {2011})}\BibitemShut {NoStop}%
\bibitem [{\citenamefont {Hsu}\ \emph {et~al.}(2011)\citenamefont {Hsu}, \citenamefont {Kraemer}, \citenamefont {Oliver}, \citenamefont {Schlichting},\ and\ \citenamefont {Thompson-Schill}}]{HsuEtAl2011}%
  \BibitemOpen
  \bibfield  {author} {\bibinfo {author} {\bibfnamefont {N.~S.}\ \bibnamefont {Hsu}}, \bibinfo {author} {\bibfnamefont {D.~J.~M.}\ \bibnamefont {Kraemer}}, \bibinfo {author} {\bibfnamefont {R.~T.}\ \bibnamefont {Oliver}}, \bibinfo {author} {\bibfnamefont {M.~L.}\ \bibnamefont {Schlichting}},\ and\ \bibinfo {author} {\bibfnamefont {S.~L.}\ \bibnamefont {Thompson-Schill}},\ }\bibfield  {title} {\bibinfo {title} {{Color, Context, and Cognitive Style: Variations in Color Knowledge Retrieval as a Function of Task and Subject Variables}},\ }\href {https://doi.org/10.1162/jocn.2011.21619} {\bibfield  {journal} {\bibinfo  {journal} {Journal of Cognitive Neuroscience}\ }\textbf {\bibinfo {volume} {23}},\ \bibinfo {pages} {2544} (\bibinfo {year} {2011})}\BibitemShut {NoStop}%
\bibitem [{\citenamefont {Athanasopoulos}(2009)}]{Athanasopoulos2009}%
  \BibitemOpen
  \bibfield  {author} {\bibinfo {author} {\bibfnamefont {P.}~\bibnamefont {Athanasopoulos}},\ }\bibfield  {title} {\bibinfo {title} {{Cognitive representation of colour in bilinguals: The case of Greek blues}},\ }\href {https://doi.org/10.1017/S136672890800388X} {\bibfield  {journal} {\bibinfo  {journal} {Bilingualism: Language and Cognition}\ }\textbf {\bibinfo {volume} {12}},\ \bibinfo {pages} {83} (\bibinfo {year} {2009})}\BibitemShut {NoStop}%
\bibitem [{\citenamefont {Thompson}\ \emph {et~al.}(2020)\citenamefont {Thompson}, \citenamefont {Roberts},\ and\ \citenamefont {Lupyan}}]{ThompsonEtAl2020}%
  \BibitemOpen
  \bibfield  {author} {\bibinfo {author} {\bibfnamefont {B.}~\bibnamefont {Thompson}}, \bibinfo {author} {\bibfnamefont {S.~G.}\ \bibnamefont {Roberts}},\ and\ \bibinfo {author} {\bibfnamefont {G.}~\bibnamefont {Lupyan}},\ }\bibfield  {title} {\bibinfo {title} {{Cultural influences on word meanings revealed through large-scale semantic alignment}},\ }\href {https://doi.org/10.1038/s41562-020-0924-8} {\bibfield  {journal} {\bibinfo  {journal} {Nature Human Behaviour}\ }\textbf {\bibinfo {volume} {4}},\ \bibinfo {pages} {1029} (\bibinfo {year} {2020})}\BibitemShut {NoStop}%
\bibitem [{\citenamefont {Johns}(2021)}]{Johns2021}%
  \BibitemOpen
  \bibfield  {author} {\bibinfo {author} {\bibfnamefont {B.~T.}\ \bibnamefont {Johns}},\ }\bibfield  {title} {\bibinfo {title} {{Distributional social semantics: Inferring word meanings from communication patterns}},\ }\href {https://doi.org/10.1016/j.cogpsych.2021.101441} {\bibfield  {journal} {\bibinfo  {journal} {Cognitive Psychology}\ }\textbf {\bibinfo {volume} {131}},\ \bibinfo {pages} {101441} (\bibinfo {year} {2021})}\BibitemShut {NoStop}%
\bibitem [{\citenamefont {Cesario}\ \emph {et~al.}(2010)\citenamefont {Cesario}, \citenamefont {Plaks}, \citenamefont {Hagiwara}, \citenamefont {Navarrete},\ and\ \citenamefont {Higgins}}]{CesarioEtAl2010}%
  \BibitemOpen
  \bibfield  {author} {\bibinfo {author} {\bibfnamefont {J.}~\bibnamefont {Cesario}}, \bibinfo {author} {\bibfnamefont {J.~E.}\ \bibnamefont {Plaks}}, \bibinfo {author} {\bibfnamefont {N.}~\bibnamefont {Hagiwara}}, \bibinfo {author} {\bibfnamefont {C.~D.}\ \bibnamefont {Navarrete}},\ and\ \bibinfo {author} {\bibfnamefont {E.~T.}\ \bibnamefont {Higgins}},\ }\bibfield  {title} {\bibinfo {title} {{The Ecology of Automaticity: How Situational Contingencies Shape Action Semantics and Social Behavior}},\ }\href {https://doi.org/10.1177/0956797610378685} {\bibfield  {journal} {\bibinfo  {journal} {Psychological Science}\ }\textbf {\bibinfo {volume} {21}},\ \bibinfo {pages} {1311} (\bibinfo {year} {2010})}\BibitemShut {NoStop}%
\bibitem [{\citenamefont {Marian}\ and\ \citenamefont {Kaushanskaya}(2007)}]{MarianKaushanskaya2007}%
  \BibitemOpen
  \bibfield  {author} {\bibinfo {author} {\bibfnamefont {V.}~\bibnamefont {Marian}}\ and\ \bibinfo {author} {\bibfnamefont {M.}~\bibnamefont {Kaushanskaya}},\ }\bibfield  {title} {\bibinfo {title} {{Language context guides memory content}},\ }\href {https://doi.org/10.3758/BF03194123} {\bibfield  {journal} {\bibinfo  {journal} {Psychonomic Bulletin \& Review}\ }\textbf {\bibinfo {volume} {14}},\ \bibinfo {pages} {925} (\bibinfo {year} {2007})}\BibitemShut {NoStop}%
\bibitem [{\citenamefont {Barsalou}(2009)}]{Barsalou2009}%
  \BibitemOpen
  \bibfield  {author} {\bibinfo {author} {\bibfnamefont {L.~W.}\ \bibnamefont {Barsalou}},\ }\bibfield  {title} {\bibinfo {title} {{Simulation, situated conceptualization, and prediction}},\ }\href {https://doi.org/10.1098/rstb.2008.0319} {\bibfield  {journal} {\bibinfo  {journal} {Philosophical Transactions of the Royal Society B: Biological Sciences}\ }\textbf {\bibinfo {volume} {364}},\ \bibinfo {pages} {1281} (\bibinfo {year} {2009})}\BibitemShut {NoStop}%
\bibitem [{\citenamefont {Pecher}\ \emph {et~al.}(1998)\citenamefont {Pecher}, \citenamefont {Zeelenberg},\ and\ \citenamefont {Raaijmakers}}]{PecherEtAl1998}%
  \BibitemOpen
  \bibfield  {author} {\bibinfo {author} {\bibfnamefont {D.}~\bibnamefont {Pecher}}, \bibinfo {author} {\bibfnamefont {R.}~\bibnamefont {Zeelenberg}},\ and\ \bibinfo {author} {\bibfnamefont {J.~G.~W.}\ \bibnamefont {Raaijmakers}},\ }\bibfield  {title} {\bibinfo {title} {{Does Pizza Prime Coin? Perceptual Priming in Lexical Decision and Pronunciation}},\ }\href {https://doi.org/10.1006/jmla.1997.2557} {\bibfield  {journal} {\bibinfo  {journal} {Journal of Memory and Language}\ }\textbf {\bibinfo {volume} {38}},\ \bibinfo {pages} {401} (\bibinfo {year} {1998})}\BibitemShut {NoStop}%
\bibitem [{\citenamefont {Borghi}\ \emph {et~al.}(2004)\citenamefont {Borghi}, \citenamefont {Glenberg},\ and\ \citenamefont {Kaschak}}]{BorghiEtAl2004}%
  \BibitemOpen
  \bibfield  {author} {\bibinfo {author} {\bibfnamefont {A.~M.}\ \bibnamefont {Borghi}}, \bibinfo {author} {\bibfnamefont {A.~M.}\ \bibnamefont {Glenberg}},\ and\ \bibinfo {author} {\bibfnamefont {M.~P.}\ \bibnamefont {Kaschak}},\ }\bibfield  {title} {\bibinfo {title} {{Putting words in perspective}},\ }\href {https://doi.org/10.3758/BF03196865} {\bibfield  {journal} {\bibinfo  {journal} {Memory \& Cognition}\ }\textbf {\bibinfo {volume} {32}},\ \bibinfo {pages} {863} (\bibinfo {year} {2004})}\BibitemShut {NoStop}%
\bibitem [{\citenamefont {Connell}\ and\ \citenamefont {Lynott}(2014)}]{ConnellLynott2014}%
  \BibitemOpen
  \bibfield  {author} {\bibinfo {author} {\bibfnamefont {L.}~\bibnamefont {Connell}}\ and\ \bibinfo {author} {\bibfnamefont {D.}~\bibnamefont {Lynott}},\ }\bibfield  {title} {\bibinfo {title} {{Principles of Representation: Why You Can't Represent the Same Concept Twice}},\ }\href {https://doi.org/10.1111/tops.12097} {\bibfield  {journal} {\bibinfo  {journal} {Topics in Cognitive Science}\ }\textbf {\bibinfo {volume} {6}},\ \bibinfo {pages} {390} (\bibinfo {year} {2014})}\BibitemShut {NoStop}%
\bibitem [{\citenamefont {Musz}\ and\ \citenamefont {Thompson-Schill}(2015)}]{MuszThompsonSchill2015}%
  \BibitemOpen
  \bibfield  {author} {\bibinfo {author} {\bibfnamefont {E.}~\bibnamefont {Musz}}\ and\ \bibinfo {author} {\bibfnamefont {S.~L.}\ \bibnamefont {Thompson-Schill}},\ }\bibfield  {title} {\bibinfo {title} {{Semantic variability predicts neural variability of object concepts}},\ }\href {https://doi.org/10.1016/j.neuropsychologia.2014.11.029} {\bibfield  {journal} {\bibinfo  {journal} {Neuropsychologia}\ }\textbf {\bibinfo {volume} {76}},\ \bibinfo {pages} {41} (\bibinfo {year} {2015})}\BibitemShut {NoStop}%
\bibitem [{\citenamefont {Vervaeke}\ and\ \citenamefont {Ferraro}(2013{\natexlab{a}})}]{VervaekeFerraro2013Relevance}%
  \BibitemOpen
  \bibfield  {author} {\bibinfo {author} {\bibfnamefont {J.}~\bibnamefont {Vervaeke}}\ and\ \bibinfo {author} {\bibfnamefont {L.}~\bibnamefont {Ferraro}},\ }\bibfield  {title} {\bibinfo {title} {Relevance, meaning and the cognitive science of wisdom}\ }(\bibinfo {year} {2013})\BibitemShut {NoStop}%
\bibitem [{\citenamefont {Andersen}\ \emph {et~al.}(ming)\citenamefont {Andersen}, \citenamefont {Miller},\ and\ \citenamefont {Vervaeke}}]{andersen2022rr}%
  \BibitemOpen
  \bibfield  {author} {\bibinfo {author} {\bibfnamefont {B.~P.}\ \bibnamefont {Andersen}}, \bibinfo {author} {\bibfnamefont {M.}~\bibnamefont {Miller}},\ and\ \bibinfo {author} {\bibfnamefont {J.}~\bibnamefont {Vervaeke}},\ }\bibfield  {title} {\bibinfo {title} {Predictive processing and relevance realization: Exploring convergent solutions to the frame problem},\ }\href {https://doi.org/10.1007/s11097-022-09850-6} {\bibfield  {journal} {\bibinfo  {journal} {Phenomenology and the Cognitive Sciences}\ ,\ \bibinfo {pages} {1}} (\bibinfo {year} {forthcoming})}\BibitemShut {NoStop}%
\bibitem [{\citenamefont {Jaeger}\ \emph {et~al.}(2023)\citenamefont {Jaeger}, \citenamefont {Riedl}, \citenamefont {Djedovic}, \citenamefont {Vervaeke},\ and\ \citenamefont {Walsh}}]{jaeger_riedl_djedovic_vervaeke_walsh_2023}%
  \BibitemOpen
  \bibfield  {author} {\bibinfo {author} {\bibfnamefont {J.}~\bibnamefont {Jaeger}}, \bibinfo {author} {\bibfnamefont {A.}~\bibnamefont {Riedl}}, \bibinfo {author} {\bibfnamefont {A.}~\bibnamefont {Djedovic}}, \bibinfo {author} {\bibfnamefont {J.}~\bibnamefont {Vervaeke}},\ and\ \bibinfo {author} {\bibfnamefont {D.}~\bibnamefont {Walsh}},\ }\href {https://doi.org/10.3389/fpsyg.2024.1362658} {\bibinfo {title} {Naturalizing relevance realization: Why agency and cognition are fundamentally not computational}} (\bibinfo {year} {2023})\BibitemShut {NoStop}%
\bibitem [{\citenamefont {Vervaeke}\ and\ \citenamefont {Ferraro}(2013{\natexlab{b}})}]{VervaekeRobinson2013Cognitive}%
  \BibitemOpen
  \bibfield  {author} {\bibinfo {author} {\bibfnamefont {J.}~\bibnamefont {Vervaeke}}\ and\ \bibinfo {author} {\bibfnamefont {L.}~\bibnamefont {Ferraro}},\ }\bibfield  {title} {\bibinfo {title} {Relevance realization and the neurodynamics and neuroconnectivity of general intelligence},\ }in\ \href@noop {} {\emph {\bibinfo {booktitle} {SmartData}}},\ \bibinfo {editor} {edited by\ \bibinfo {editor} {\bibfnamefont {I.}~\bibnamefont {Harvey}}, \bibinfo {editor} {\bibfnamefont {A.}~\bibnamefont {Cavoukian}}, \bibinfo {editor} {\bibfnamefont {G.}~\bibnamefont {Tomko}}, \bibinfo {editor} {\bibfnamefont {D.}~\bibnamefont {Borrett}}, \bibinfo {editor} {\bibfnamefont {H.}~\bibnamefont {Kwan}},\ and\ \bibinfo {editor} {\bibfnamefont {D.}~\bibnamefont {Hatzinakos}}}\ (\bibinfo  {publisher} {Springer New York},\ \bibinfo {address} {New York, NY},\ \bibinfo {year} {2013})\ pp.\ \bibinfo {pages} {57--68}\BibitemShut {NoStop}%
\bibitem [{\citenamefont {Hintzman}(1984)}]{Hintzman1984}%
  \BibitemOpen
  \bibfield  {author} {\bibinfo {author} {\bibfnamefont {D.~L.}\ \bibnamefont {Hintzman}},\ }\bibfield  {title} {\bibinfo {title} {{MINERVA 2: A simulation model of human memory}},\ }\href@noop {} {\bibfield  {journal} {\bibinfo  {journal} {Behavior Research Methods, Instruments, \& Computers}\ }\textbf {\bibinfo {volume} {16}},\ \bibinfo {pages} {96} (\bibinfo {year} {1984})}\BibitemShut {NoStop}%
\bibitem [{\citenamefont {Jamieson}\ \emph {et~al.}(2018)\citenamefont {Jamieson}, \citenamefont {Avery}, \citenamefont {Johns},\ and\ \citenamefont {Jones}}]{JamiesonEtAl2018}%
  \BibitemOpen
  \bibfield  {author} {\bibinfo {author} {\bibfnamefont {R.~K.}\ \bibnamefont {Jamieson}}, \bibinfo {author} {\bibfnamefont {J.~E.}\ \bibnamefont {Avery}}, \bibinfo {author} {\bibfnamefont {B.~T.}\ \bibnamefont {Johns}},\ and\ \bibinfo {author} {\bibfnamefont {M.~N.}\ \bibnamefont {Jones}},\ }\bibfield  {title} {\bibinfo {title} {{An Instance Theory of Semantic Memory}},\ }\href {https://doi.org/10.1007/s42113-018-0008-2} {\bibfield  {journal} {\bibinfo  {journal} {Computational Brain \& Behavior}\ }\textbf {\bibinfo {volume} {1}},\ \bibinfo {pages} {119} (\bibinfo {year} {2018})}\BibitemShut {NoStop}%
\bibitem [{\citenamefont {Van~Dam}\ \emph {et~al.}(2010)\citenamefont {Van~Dam}, \citenamefont {Rueschemeyer}, \citenamefont {Lindemann},\ and\ \citenamefont {Bekkering}}]{VanDamEtAl2010}%
  \BibitemOpen
  \bibfield  {author} {\bibinfo {author} {\bibfnamefont {W.~O.}\ \bibnamefont {Van~Dam}}, \bibinfo {author} {\bibfnamefont {S.-A.}\ \bibnamefont {Rueschemeyer}}, \bibinfo {author} {\bibfnamefont {O.}~\bibnamefont {Lindemann}},\ and\ \bibinfo {author} {\bibfnamefont {H.}~\bibnamefont {Bekkering}},\ }\bibfield  {title} {\bibinfo {title} {{Context Effects in Embodied Lexical-Semantic Processing}},\ }\href {https://doi.org/10.3389/fpsyg.2010.00150} {\bibfield  {journal} {\bibinfo  {journal} {Frontiers in Psychology}\ }\textbf {\bibinfo {volume} {1}},\ \bibinfo {pages} {150} (\bibinfo {year} {2010})}\BibitemShut {NoStop}%
\bibitem [{\citenamefont {Van~Dam}\ \emph {et~al.}(2012)\citenamefont {Van~Dam}, \citenamefont {Van~Dijk}, \citenamefont {Bekkering},\ and\ \citenamefont {Rueschemeyer}}]{VanDamEtAl2012}%
  \BibitemOpen
  \bibfield  {author} {\bibinfo {author} {\bibfnamefont {W.~O.}\ \bibnamefont {Van~Dam}}, \bibinfo {author} {\bibfnamefont {M.}~\bibnamefont {Van~Dijk}}, \bibinfo {author} {\bibfnamefont {H.}~\bibnamefont {Bekkering}},\ and\ \bibinfo {author} {\bibfnamefont {S.-A.}\ \bibnamefont {Rueschemeyer}},\ }\bibfield  {title} {\bibinfo {title} {{Flexibility in embodied lexical-semantic representations}},\ }\href@noop {} {\bibfield  {journal} {\bibinfo  {journal} {Human Brain Mapping}\ }\textbf {\bibinfo {volume} {33}},\ \bibinfo {pages} {2322} (\bibinfo {year} {2012})}\BibitemShut {NoStop}%
\bibitem [{\citenamefont {Aerts}(2009)}]{Aerts2009}%
  \BibitemOpen
  \bibfield  {author} {\bibinfo {author} {\bibfnamefont {D.}~\bibnamefont {Aerts}},\ }\bibfield  {title} {\bibinfo {title} {Quantum structure in cognition},\ }\href {https://doi.org/https://doi.org/10.1016/j.jmp.2009.04.005} {\bibfield  {journal} {\bibinfo  {journal} {Journal of Mathematical Psychology}\ }\textbf {\bibinfo {volume} {53}},\ \bibinfo {pages} {314} (\bibinfo {year} {2009})},\ \bibinfo {note} {special Issue: Quantum Cognition}\BibitemShut {NoStop}%
\bibitem [{\citenamefont {Bruza}\ \emph {et~al.}(2009)\citenamefont {Bruza}, \citenamefont {Busemeyer},\ and\ \citenamefont {Gabora}}]{Bruza2009}%
  \BibitemOpen
  \bibfield  {author} {\bibinfo {author} {\bibfnamefont {P.}~\bibnamefont {Bruza}}, \bibinfo {author} {\bibfnamefont {J.~R.}\ \bibnamefont {Busemeyer}},\ and\ \bibinfo {author} {\bibfnamefont {L.}~\bibnamefont {Gabora}},\ }\bibfield  {title} {\bibinfo {title} {Introduction to the special issue on quantum cognition},\ }\href {https://doi.org/https://doi.org/10.1016/j.jmp.2009.06.002} {\bibfield  {journal} {\bibinfo  {journal} {Journal of Mathematical Psychology}\ }\textbf {\bibinfo {volume} {53}},\ \bibinfo {pages} {303} (\bibinfo {year} {2009})},\ \bibinfo {note} {special Issue: Quantum Cognition}\BibitemShut {NoStop}%
\bibitem [{\citenamefont {Gabora}\ and\ \citenamefont {and}(2002)}]{Gabora2002}%
  \BibitemOpen
  \bibfield  {author} {\bibinfo {author} {\bibfnamefont {L.}~\bibnamefont {Gabora}}\ and\ \bibinfo {author} {\bibfnamefont {D.~A.}\ \bibnamefont {and}},\ }\bibfield  {title} {\bibinfo {title} {Contextualizing concepts using a mathematical generalization of the quantum formalism},\ }\href {https://doi.org/10.1080/09528130210162253} {\bibfield  {journal} {\bibinfo  {journal} {Journal of Experimental \& Theoretical Artificial Intelligence}\ }\textbf {\bibinfo {volume} {14}},\ \bibinfo {pages} {327} (\bibinfo {year} {2002})},\ \Eprint {https://arxiv.org/abs/https://doi.org/10.1080/09528130210162253} {https://doi.org/10.1080/09528130210162253} \BibitemShut {NoStop}%
\bibitem [{\citenamefont {Bruza}\ \emph {et~al.}(2012)\citenamefont {Bruza}, \citenamefont {Kitto}, \citenamefont {Ramm}, \citenamefont {Sitbon},\ and\ \citenamefont {Song}}]{bruza2012}%
  \BibitemOpen
  \bibfield  {author} {\bibinfo {author} {\bibfnamefont {P.}~\bibnamefont {Bruza}}, \bibinfo {author} {\bibfnamefont {K.}~\bibnamefont {Kitto}}, \bibinfo {author} {\bibfnamefont {B.}~\bibnamefont {Ramm}}, \bibinfo {author} {\bibfnamefont {L.}~\bibnamefont {Sitbon}},\ and\ \bibinfo {author} {\bibfnamefont {D.}~\bibnamefont {Song}},\ }\bibfield  {title} {\bibinfo {title} {Quantum-like non-separability of concept combinations, emergent associates and abduction},\ }\href {https://doi.org/10.1093/jigpal/jzq049} {\bibfield  {journal} {\bibinfo  {journal} {Logic Journal of the IGPL}\ }\textbf {\bibinfo {volume} {20}},\ \bibinfo {pages} {445} (\bibinfo {year} {2012})}\BibitemShut {NoStop}%
\bibitem [{\citenamefont {Bruza}\ \emph {et~al.}(2015)\citenamefont {Bruza}, \citenamefont {Kitto}, \citenamefont {Ramm},\ and\ \citenamefont {Sitbon}}]{bruza2015}%
  \BibitemOpen
  \bibfield  {author} {\bibinfo {author} {\bibfnamefont {P.}~\bibnamefont {Bruza}}, \bibinfo {author} {\bibfnamefont {K.}~\bibnamefont {Kitto}}, \bibinfo {author} {\bibfnamefont {B.}~\bibnamefont {Ramm}},\ and\ \bibinfo {author} {\bibfnamefont {L.}~\bibnamefont {Sitbon}},\ }\bibfield  {title} {\bibinfo {title} {A probabilistic framework for analysing the compositionality of conceptual combinations},\ }\href@noop {} {\bibfield  {journal} {\bibinfo  {journal} {Journal of Mathematical Psychology}\ }\textbf {\bibinfo {volume} {67}},\ \bibinfo {pages} {26} (\bibinfo {year} {2015})}\BibitemShut {NoStop}%
\bibitem [{\citenamefont {Yearsley}\ and\ \citenamefont {Busemeyer}(2016)}]{Yearsley2016}%
  \BibitemOpen
  \bibfield  {author} {\bibinfo {author} {\bibfnamefont {J.~M.}\ \bibnamefont {Yearsley}}\ and\ \bibinfo {author} {\bibfnamefont {J.~R.}\ \bibnamefont {Busemeyer}},\ }\bibfield  {title} {\bibinfo {title} {Quantum cognition and decision theories: A tutorial},\ }\href {https://doi.org/https://doi.org/10.1016/j.jmp.2015.11.005} {\bibfield  {journal} {\bibinfo  {journal} {Journal of Mathematical Psychology}\ }\textbf {\bibinfo {volume} {74}},\ \bibinfo {pages} {99} (\bibinfo {year} {2016})},\ \bibinfo {note} {foundations of Probability Theory in Psychology and Beyond}\BibitemShut {NoStop}%
\bibitem [{\citenamefont {Pothos}\ and\ \citenamefont {Busemeyer}(2022)}]{pothos2022}%
  \BibitemOpen
  \bibfield  {author} {\bibinfo {author} {\bibfnamefont {E.~M.}\ \bibnamefont {Pothos}}\ and\ \bibinfo {author} {\bibfnamefont {J.~R.}\ \bibnamefont {Busemeyer}},\ }\bibfield  {title} {\bibinfo {title} {Quantum cognition},\ }\href {https://doi.org/https://doi.org/10.1146/annurev-psych-033020-123501} {\bibfield  {journal} {\bibinfo  {journal} {Annual Review of Psychology}\ }\textbf {\bibinfo {volume} {73}},\ \bibinfo {pages} {749} (\bibinfo {year} {2022})}\BibitemShut {NoStop}%
\bibitem [{\citenamefont {Aerts}\ \emph {et~al.}(2010)\citenamefont {Aerts}, \citenamefont {Czachor}, \citenamefont {D'Hooghe}, \citenamefont {Sozzo} \emph {et~al.}}]{aerts2010pet}%
  \BibitemOpen
  \bibfield  {author} {\bibinfo {author} {\bibfnamefont {D.}~\bibnamefont {Aerts}}, \bibinfo {author} {\bibfnamefont {M.}~\bibnamefont {Czachor}}, \bibinfo {author} {\bibfnamefont {B.}~\bibnamefont {D'Hooghe}}, \bibinfo {author} {\bibfnamefont {S.}~\bibnamefont {Sozzo}}, \emph {et~al.},\ }\bibfield  {title} {\bibinfo {title} {The pet-fish problem on the world-wide web.},\ }in\ \href@noop {} {\emph {\bibinfo {booktitle} {AAAI Fall Symposium: Quantum Informatics for Cognitive, Social, and Semantic Processes}}}\ (\bibinfo {year} {2010})\BibitemShut {NoStop}%
\bibitem [{\citenamefont {Bell}(1964)}]{bell1964}%
  \BibitemOpen
  \bibfield  {author} {\bibinfo {author} {\bibfnamefont {J.~S.}\ \bibnamefont {Bell}},\ }\bibfield  {title} {\bibinfo {title} {On the einstein podolsky rosen paradox},\ }\href {https://doi.org/10.1103/PhysicsPhysiqueFizika.1.195} {\bibfield  {journal} {\bibinfo  {journal} {Physics Physique Fizika}\ }\textbf {\bibinfo {volume} {1}},\ \bibinfo {pages} {195} (\bibinfo {year} {1964})}\BibitemShut {NoStop}%
\bibitem [{\citenamefont {Uprety}(2020)}]{uprety2020investigation}%
  \BibitemOpen
  \bibfield  {author} {\bibinfo {author} {\bibfnamefont {S.}~\bibnamefont {Uprety}},\ }\href@noop {} {\bibinfo {title} {Investigation and modelling of quantum-like user cognitive behaviour in information access and retrieval}} (\bibinfo {year} {2020})\BibitemShut {NoStop}%
\bibitem [{\citenamefont {Aerts}\ \emph {et~al.}(2000)\citenamefont {Aerts}, \citenamefont {Aerts}, \citenamefont {Broekaert},\ and\ \citenamefont {Gabora}}]{aerts2000}%
  \BibitemOpen
  \bibfield  {author} {\bibinfo {author} {\bibfnamefont {D.}~\bibnamefont {Aerts}}, \bibinfo {author} {\bibfnamefont {S.}~\bibnamefont {Aerts}}, \bibinfo {author} {\bibfnamefont {J.}~\bibnamefont {Broekaert}},\ and\ \bibinfo {author} {\bibfnamefont {L.}~\bibnamefont {Gabora}},\ }\bibfield  {title} {\bibinfo {title} {The violation of bell inequalities in the macroworld},\ }\href {https://doi.org/10.1023/a:1026449716544} {\bibfield  {journal} {\bibinfo  {journal} {Foundations of Physics}\ }\textbf {\bibinfo {volume} {30}},\ \bibinfo {pages} {1387} (\bibinfo {year} {2000})}\BibitemShut {NoStop}%
\bibitem [{\citenamefont {Aerts}\ \emph {et~al.}(2018{\natexlab{a}})\citenamefont {Aerts}, \citenamefont {Argu\"{e}lles}, \citenamefont {Beltran}, \citenamefont {Geriente}, \citenamefont {de~Bianchi}, \citenamefont {Sozzo},\ and\ \citenamefont {Veloz}}]{aerts2018_wind1}%
  \BibitemOpen
  \bibfield  {author} {\bibinfo {author} {\bibfnamefont {D.}~\bibnamefont {Aerts}}, \bibinfo {author} {\bibfnamefont {J.~A.}\ \bibnamefont {Argu\"{e}lles}}, \bibinfo {author} {\bibfnamefont {L.}~\bibnamefont {Beltran}}, \bibinfo {author} {\bibfnamefont {S.}~\bibnamefont {Geriente}}, \bibinfo {author} {\bibfnamefont {M.~S.}\ \bibnamefont {de~Bianchi}}, \bibinfo {author} {\bibfnamefont {S.}~\bibnamefont {Sozzo}},\ and\ \bibinfo {author} {\bibfnamefont {T.}~\bibnamefont {Veloz}},\ }\bibfield  {title} {\bibinfo {title} {Spin and wind directions i: Identifying entanglement in nature and cognition},\ }\href {https://doi.org/10.1007/s10699-017-9528-9} {\bibfield  {journal} {\bibinfo  {journal} {Foundations of Science}\ }\textbf {\bibinfo {volume} {23}},\ \bibinfo {pages} {323} (\bibinfo {year} {2018}{\natexlab{a}})}\BibitemShut {NoStop}%
\bibitem [{\citenamefont {Aerts}\ \emph {et~al.}(2018{\natexlab{b}})\citenamefont {Aerts}, \citenamefont {Argu{\"e}lles}, \citenamefont {Beltran}, \citenamefont {Geriente}, \citenamefont {Sassoli~de Bianchi}, \citenamefont {Sozzo},\ and\ \citenamefont {Veloz}}]{aerts2018_wind2}%
  \BibitemOpen
  \bibfield  {author} {\bibinfo {author} {\bibfnamefont {D.}~\bibnamefont {Aerts}}, \bibinfo {author} {\bibfnamefont {J.~A.}\ \bibnamefont {Argu{\"e}lles}}, \bibinfo {author} {\bibfnamefont {L.}~\bibnamefont {Beltran}}, \bibinfo {author} {\bibfnamefont {S.}~\bibnamefont {Geriente}}, \bibinfo {author} {\bibfnamefont {M.}~\bibnamefont {Sassoli~de Bianchi}}, \bibinfo {author} {\bibfnamefont {S.}~\bibnamefont {Sozzo}},\ and\ \bibinfo {author} {\bibfnamefont {T.}~\bibnamefont {Veloz}},\ }\bibfield  {title} {\bibinfo {title} {Spin and wind directions ii: A bell state quantum model},\ }\href@noop {} {\bibfield  {journal} {\bibinfo  {journal} {Foundations of Science}\ }\textbf {\bibinfo {volume} {23}},\ \bibinfo {pages} {337} (\bibinfo {year} {2018}{\natexlab{b}})}\BibitemShut {NoStop}%
\bibitem [{\citenamefont {Bruza}\ \emph {et~al.}(2023)\citenamefont {Bruza}, \citenamefont {Fell}, \citenamefont {Hoyte}, \citenamefont {Dehdashti}, \citenamefont {Obeid}, \citenamefont {Gibson},\ and\ \citenamefont {Moreira}}]{bruza2023}%
  \BibitemOpen
  \bibfield  {author} {\bibinfo {author} {\bibfnamefont {P.}~\bibnamefont {Bruza}}, \bibinfo {author} {\bibfnamefont {L.}~\bibnamefont {Fell}}, \bibinfo {author} {\bibfnamefont {P.}~\bibnamefont {Hoyte}}, \bibinfo {author} {\bibfnamefont {S.}~\bibnamefont {Dehdashti}}, \bibinfo {author} {\bibfnamefont {A.}~\bibnamefont {Obeid}}, \bibinfo {author} {\bibfnamefont {A.}~\bibnamefont {Gibson}},\ and\ \bibinfo {author} {\bibfnamefont {C.}~\bibnamefont {Moreira}},\ }\bibfield  {title} {\bibinfo {title} {Contextuality and context-sensitivity in probabilistic models of cognition},\ }\href {https://doi.org/https://doi.org/10.1016/j.cogpsych.2022.101529} {\bibfield  {journal} {\bibinfo  {journal} {Cognitive Psychology}\ }\textbf {\bibinfo {volume} {140}},\ \bibinfo {pages} {101529} (\bibinfo {year} {2023})}\BibitemShut {NoStop}%
\bibitem [{\citenamefont {Kolmogorov}(1965)}]{Kolmogorov65}%
  \BibitemOpen
  \bibfield  {author} {\bibinfo {author} {\bibfnamefont {A.~N.}\ \bibnamefont {Kolmogorov}},\ }\bibfield  {title} {\bibinfo {title} {Three approaches to the quantitative definition of information},\ }\href@noop {} {\bibfield  {journal} {\bibinfo  {journal} {Problems of Information Transmission}\ }\textbf {\bibinfo {volume} {1}},\ \bibinfo {pages} {1} (\bibinfo {year} {1965})}\BibitemShut {NoStop}%
\bibitem [{\citenamefont {Clauser}\ \emph {et~al.}(1969)\citenamefont {Clauser}, \citenamefont {Horne}, \citenamefont {Shimony},\ and\ \citenamefont {Holt}}]{clauser1969}%
  \BibitemOpen
  \bibfield  {author} {\bibinfo {author} {\bibfnamefont {J.~F.}\ \bibnamefont {Clauser}}, \bibinfo {author} {\bibfnamefont {M.~A.}\ \bibnamefont {Horne}}, \bibinfo {author} {\bibfnamefont {A.}~\bibnamefont {Shimony}},\ and\ \bibinfo {author} {\bibfnamefont {R.~A.}\ \bibnamefont {Holt}},\ }\bibfield  {title} {\bibinfo {title} {Proposed experiment to test local hidden-variable theories},\ }\href {https://doi.org/10.1103/PhysRevLett.23.880} {\bibfield  {journal} {\bibinfo  {journal} {Phys. Rev. Lett.}\ }\textbf {\bibinfo {volume} {23}},\ \bibinfo {pages} {880} (\bibinfo {year} {1969})}\BibitemShut {NoStop}%
\bibitem [{\citenamefont {Guo}\ and\ \citenamefont {Caliskan}(2021)}]{GuoCaliskan2021}%
  \BibitemOpen
  \bibfield  {author} {\bibinfo {author} {\bibfnamefont {W.}~\bibnamefont {Guo}}\ and\ \bibinfo {author} {\bibfnamefont {A.}~\bibnamefont {Caliskan}},\ }\bibfield  {title} {\bibinfo {title} {{Detecting Emergent Intersectional Biases: Contextualized Word Embeddings Contain a Distribution of Human-like Biases}},\ }in\ \href {https://doi.org/10.1145/3461702.3462536} {\emph {\bibinfo {booktitle} {Proceedings of the 2021 AAAI/ACM Conference on AI, Ethics, and Society}}}\ (\bibinfo {year} {2021})\ pp.\ \bibinfo {pages} {122--133}\BibitemShut {NoStop}%
\bibitem [{\citenamefont {Payne}\ \emph {et~al.}(2017)\citenamefont {Payne}, \citenamefont {Vuletich},\ and\ \citenamefont {Lundberg}}]{PayneEtAl2017}%
  \BibitemOpen
  \bibfield  {author} {\bibinfo {author} {\bibfnamefont {B.~K.}\ \bibnamefont {Payne}}, \bibinfo {author} {\bibfnamefont {H.~A.}\ \bibnamefont {Vuletich}},\ and\ \bibinfo {author} {\bibfnamefont {K.~B.}\ \bibnamefont {Lundberg}},\ }\bibfield  {title} {\bibinfo {title} {{The Bias of Crowds: How Implicit Bias Bridges Personal and Systemic Prejudice}},\ }\href {https://doi.org/10.1080/1047840X.2017.1335568} {\bibfield  {journal} {\bibinfo  {journal} {Psychological Inquiry}\ }\textbf {\bibinfo {volume} {28}},\ \bibinfo {pages} {233} (\bibinfo {year} {2017})}\BibitemShut {NoStop}%
\bibitem [{\citenamefont {Siddique}\ \emph {et~al.}(2025)\citenamefont {Siddique}, \citenamefont {Khalid}, \citenamefont {Turner},\ and\ \citenamefont {Espinosa-Anke}}]{siddique2025}%
  \BibitemOpen
  \bibfield  {author} {\bibinfo {author} {\bibfnamefont {Z.}~\bibnamefont {Siddique}}, \bibinfo {author} {\bibfnamefont {I.}~\bibnamefont {Khalid}}, \bibinfo {author} {\bibfnamefont {L.~D.}\ \bibnamefont {Turner}},\ and\ \bibinfo {author} {\bibfnamefont {L.}~\bibnamefont {Espinosa-Anke}},\ }\bibfield  {title} {\bibinfo {title} {Shifting perspectives: Steering vector ensembles for robust bias mitigation in llms},\ }\href {https://arxiv.org/abs/2503.05371} {\bibfield  {journal} {\bibinfo  {journal} {arXiv preprint arXiv:2503.05371}\ } (\bibinfo {year} {2025})}\BibitemShut {NoStop}%
\bibitem [{\citenamefont {Gallegos}\ \emph {et~al.}(2023)\citenamefont {Gallegos}, \citenamefont {Rossi}, \citenamefont {Barrow}, \citenamefont {Tanjim}, \citenamefont {Kim}, \citenamefont {Dernoncourt}, \citenamefont {Yu}, \citenamefont {Zhang},\ and\ \citenamefont {Ahmed}}]{biasSurvey2023}%
  \BibitemOpen
  \bibfield  {author} {\bibinfo {author} {\bibfnamefont {I.~O.}\ \bibnamefont {Gallegos}}, \bibinfo {author} {\bibfnamefont {R.~A.}\ \bibnamefont {Rossi}}, \bibinfo {author} {\bibfnamefont {J.}~\bibnamefont {Barrow}}, \bibinfo {author} {\bibfnamefont {M.~M.}\ \bibnamefont {Tanjim}}, \bibinfo {author} {\bibfnamefont {S.}~\bibnamefont {Kim}}, \bibinfo {author} {\bibfnamefont {F.}~\bibnamefont {Dernoncourt}}, \bibinfo {author} {\bibfnamefont {T.}~\bibnamefont {Yu}}, \bibinfo {author} {\bibfnamefont {R.}~\bibnamefont {Zhang}},\ and\ \bibinfo {author} {\bibfnamefont {N.~K.}\ \bibnamefont {Ahmed}},\ }\bibfield  {title} {\bibinfo {title} {Bias and fairness in large language models: A survey},\ }\href {https://arxiv.org/abs/2309.00770} {\bibfield  {journal} {\bibinfo  {journal} {arXiv preprint arXiv:2309.00770}\ } (\bibinfo {year} {2023})}\BibitemShut {NoStop}%
\bibitem [{\citenamefont {Kitadai}\ \emph {et~al.}(2024)\citenamefont {Kitadai}, \citenamefont {Ogawa},\ and\ \citenamefont {Nishino}}]{kitadai2024}%
  \BibitemOpen
  \bibfield  {author} {\bibinfo {author} {\bibfnamefont {A.}~\bibnamefont {Kitadai}}, \bibinfo {author} {\bibfnamefont {K.}~\bibnamefont {Ogawa}},\ and\ \bibinfo {author} {\bibfnamefont {N.}~\bibnamefont {Nishino}},\ }\bibfield  {title} {\bibinfo {title} {Examining the feasibility of large language models as survey respondents},\ }in\ \href {https://doi.org/10.1109/BigData62323.2024.10825497} {\emph {\bibinfo {booktitle} {2024 IEEE International Conference on Big Data (BigData)}}}\ (\bibinfo {year} {2024})\ pp.\ \bibinfo {pages} {3858--3864}\BibitemShut {NoStop}%
\bibitem [{\citenamefont {Salecha}\ \emph {et~al.}(2024)\citenamefont {Salecha}, \citenamefont {Ireland}, \citenamefont {Subrahmanya}, \citenamefont {Sedoc}, \citenamefont {Ungar},\ and\ \citenamefont {Eichstaedt}}]{salencha2024}%
  \BibitemOpen
  \bibfield  {author} {\bibinfo {author} {\bibfnamefont {A.}~\bibnamefont {Salecha}}, \bibinfo {author} {\bibfnamefont {M.~E.}\ \bibnamefont {Ireland}}, \bibinfo {author} {\bibfnamefont {S.}~\bibnamefont {Subrahmanya}}, \bibinfo {author} {\bibfnamefont {J.}~\bibnamefont {Sedoc}}, \bibinfo {author} {\bibfnamefont {L.~H.}\ \bibnamefont {Ungar}},\ and\ \bibinfo {author} {\bibfnamefont {J.~C.}\ \bibnamefont {Eichstaedt}},\ }\bibfield  {title} {\bibinfo {title} {Large language models display human-like social desirability biases in big five personality surveys},\ }\href {https://doi.org/10.1093/pnasnexus/pgae533} {\bibfield  {journal} {\bibinfo  {journal} {PNAS Nexus}\ }\textbf {\bibinfo {volume} {3}},\ \bibinfo {pages} {pgae533} (\bibinfo {year} {2024})},\ \Eprint {https://arxiv.org/abs/https://academic.oup.com/pnasnexus/article-pdf/3/12/pgae533/61188312/pgae533.pdf} {https://academic.oup.com/pnasnexus/article-pdf/3/12/pgae533/61188312/pgae533.pdf} \BibitemShut {NoStop}%
\bibitem [{\citenamefont {Tjuatja}\ \emph {et~al.}(2024)\citenamefont {Tjuatja}, \citenamefont {Chen}, \citenamefont {Wu}, \citenamefont {Talwalkwar},\ and\ \citenamefont {Neubig}}]{tjuatja2024}%
  \BibitemOpen
  \bibfield  {author} {\bibinfo {author} {\bibfnamefont {L.}~\bibnamefont {Tjuatja}}, \bibinfo {author} {\bibfnamefont {V.}~\bibnamefont {Chen}}, \bibinfo {author} {\bibfnamefont {T.}~\bibnamefont {Wu}}, \bibinfo {author} {\bibfnamefont {A.}~\bibnamefont {Talwalkwar}},\ and\ \bibinfo {author} {\bibfnamefont {G.}~\bibnamefont {Neubig}},\ }\bibfield  {title} {\bibinfo {title} {Do llms exhibit human-like response biases? a case study in survey design},\ }\href {https://doi.org/10.1162/tacl_a_00685} {\bibfield  {journal} {\bibinfo  {journal} {Transactions of the Association for Computational Linguistics}\ }\textbf {\bibinfo {volume} {12}},\ \bibinfo {pages} {1011} (\bibinfo {year} {2024})},\ \Eprint {https://arxiv.org/abs/https://direct.mit.edu/tacl/article-pdf/doi/10.1162/tacl\_a\_00685/2468689/tacl\_a\_00685.pdf} {https://direct.mit.edu/tacl/article-pdf/doi/10.1162/tacl\_a\_00685/2468689/tacl\_a\_00685.pdf} \BibitemShut {NoStop}%
\bibitem [{\citenamefont {Piantadosi}\ and\ \citenamefont {Hill}(2022)}]{piantadosi2022meaning}%
  \BibitemOpen
  \bibfield  {author} {\bibinfo {author} {\bibfnamefont {S.~T.}\ \bibnamefont {Piantadosi}}\ and\ \bibinfo {author} {\bibfnamefont {F.}~\bibnamefont {Hill}},\ }\bibfield  {title} {\bibinfo {title} {Meaning without reference in large language models},\ }\href@noop {} {\bibfield  {journal} {\bibinfo  {journal} {arXiv preprint arXiv:2208.02957}\ } (\bibinfo {year} {2022})}\BibitemShut {NoStop}%
\bibitem [{\citenamefont {Lampinen}\ \emph {et~al.}(2022)\citenamefont {Lampinen}, \citenamefont {Dasgupta}, \citenamefont {Chan}, \citenamefont {Mathewson}, \citenamefont {Tessler}, \citenamefont {Creswell}, \citenamefont {McClelland}, \citenamefont {Wang},\ and\ \citenamefont {Hill}}]{lampinen-etal-2022-language}%
  \BibitemOpen
  \bibfield  {author} {\bibinfo {author} {\bibfnamefont {A.}~\bibnamefont {Lampinen}}, \bibinfo {author} {\bibfnamefont {I.}~\bibnamefont {Dasgupta}}, \bibinfo {author} {\bibfnamefont {S.}~\bibnamefont {Chan}}, \bibinfo {author} {\bibfnamefont {K.}~\bibnamefont {Mathewson}}, \bibinfo {author} {\bibfnamefont {M.}~\bibnamefont {Tessler}}, \bibinfo {author} {\bibfnamefont {A.}~\bibnamefont {Creswell}}, \bibinfo {author} {\bibfnamefont {J.}~\bibnamefont {McClelland}}, \bibinfo {author} {\bibfnamefont {J.}~\bibnamefont {Wang}},\ and\ \bibinfo {author} {\bibfnamefont {F.}~\bibnamefont {Hill}},\ }\bibfield  {title} {\bibinfo {title} {Can language models learn from explanations in context?},\ }in\ \href {https://doi.org/10.18653/v1/2022.findings-emnlp.38} {\emph {\bibinfo {booktitle} {Findings of the Association for Computational Linguistics: EMNLP 2022}}},\ \bibinfo {editor} {edited by\ \bibinfo {editor} {\bibfnamefont {Y.}~\bibnamefont {Goldberg}}, \bibinfo {editor} {\bibfnamefont {Z.}~\bibnamefont {Kozareva}},\
  and\ \bibinfo {editor} {\bibfnamefont {Y.}~\bibnamefont {Zhang}}}\ (\bibinfo  {publisher} {Association for Computational Linguistics},\ \bibinfo {address} {Abu Dhabi, United Arab Emirates},\ \bibinfo {year} {2022})\ pp.\ \bibinfo {pages} {537--563}\BibitemShut {NoStop}%
\bibitem [{\citenamefont {Dzhafarov}\ \emph {et~al.}(2016)\citenamefont {Dzhafarov}, \citenamefont {Kujala},\ and\ \citenamefont {Cervantes}}]{dzhafarov2016contextuality}%
  \BibitemOpen
  \bibfield  {author} {\bibinfo {author} {\bibfnamefont {E.~N.}\ \bibnamefont {Dzhafarov}}, \bibinfo {author} {\bibfnamefont {J.~V.}\ \bibnamefont {Kujala}},\ and\ \bibinfo {author} {\bibfnamefont {V.~H.}\ \bibnamefont {Cervantes}},\ }\bibfield  {title} {\bibinfo {title} {Contextuality-by-default: a brief overview of ideas, concepts, and terminology},\ }in\ \href@noop {} {\emph {\bibinfo {booktitle} {Quantum Interaction: 9th International Conference, QI 2015, Filzbach, Switzerland, July 15-17, 2015, Revised Selected Papers 9}}}\ (\bibinfo {organization} {Springer},\ \bibinfo {year} {2016})\ pp.\ \bibinfo {pages} {12--23}\BibitemShut {NoStop}%
\bibitem [{\citenamefont {Cervantes}\ and\ \citenamefont {Dzhafarov}(2017)}]{cervantes2017advanced}%
  \BibitemOpen
  \bibfield  {author} {\bibinfo {author} {\bibfnamefont {V.~H.}\ \bibnamefont {Cervantes}}\ and\ \bibinfo {author} {\bibfnamefont {E.~N.}\ \bibnamefont {Dzhafarov}},\ }\bibfield  {title} {\bibinfo {title} {Advanced analysis of quantum contextuality in a psychophysical double-detection experiment},\ }\href@noop {} {\bibfield  {journal} {\bibinfo  {journal} {Journal of Mathematical Psychology}\ }\textbf {\bibinfo {volume} {79}},\ \bibinfo {pages} {77} (\bibinfo {year} {2017})}\BibitemShut {NoStop}%
\bibitem [{\citenamefont {Elman}(1990)}]{Elman1990}%
  \BibitemOpen
  \bibfield  {author} {\bibinfo {author} {\bibfnamefont {J.~L.}\ \bibnamefont {Elman}},\ }\bibfield  {title} {\bibinfo {title} {{Finding Structure in Time}},\ }\href {https://doi.org/10.1207/s15516709cog1402_1} {\bibfield  {journal} {\bibinfo  {journal} {Cognitive Science}\ }\textbf {\bibinfo {volume} {14}},\ \bibinfo {pages} {179} (\bibinfo {year} {1990})}\BibitemShut {NoStop}%
\bibitem [{\citenamefont {Wittgenstein}(1953)}]{Wittgenstein1953}%
  \BibitemOpen
  \bibfield  {author} {\bibinfo {author} {\bibfnamefont {L.}~\bibnamefont {Wittgenstein}},\ }\href@noop {} {\emph {\bibinfo {title} {Philosophical Investigations}}}\ (\bibinfo  {publisher} {Basil Blackwell},\ \bibinfo {address} {Oxford},\ \bibinfo {year} {1953})\BibitemShut {NoStop}%
\bibitem [{\citenamefont {Quine}(1960)}]{Quine1960}%
  \BibitemOpen
  \bibfield  {author} {\bibinfo {author} {\bibfnamefont {W.~V.~O.}\ \bibnamefont {Quine}},\ }\href@noop {} {\emph {\bibinfo {title} {Word \& Object}}}\ (\bibinfo  {publisher} {MIT Press},\ \bibinfo {year} {1960})\BibitemShut {NoStop}%
\bibitem [{\citenamefont {Hogarth}\ and\ \citenamefont {Einhorn}(1992)}]{Hogarth1992}%
  \BibitemOpen
  \bibfield  {author} {\bibinfo {author} {\bibfnamefont {R.~M.}\ \bibnamefont {Hogarth}}\ and\ \bibinfo {author} {\bibfnamefont {H.~J.}\ \bibnamefont {Einhorn}},\ }\bibfield  {title} {\bibinfo {title} {Order effects in belief updating: The belief-adjustment model},\ }\href {https://doi.org/https://doi.org/10.1016/0010-0285(92)90002-J} {\bibfield  {journal} {\bibinfo  {journal} {Cognitive Psychology}\ }\textbf {\bibinfo {volume} {24}},\ \bibinfo {pages} {1} (\bibinfo {year} {1992})}\BibitemShut {NoStop}%
\bibitem [{\citenamefont {Anderson}\ and\ \citenamefont {Milson}(1989)}]{AndersonMilson1989}%
  \BibitemOpen
  \bibfield  {author} {\bibinfo {author} {\bibfnamefont {J.~R.}\ \bibnamefont {Anderson}}\ and\ \bibinfo {author} {\bibfnamefont {R.}~\bibnamefont {Milson}},\ }\bibfield  {title} {\bibinfo {title} {{Human memory: An adaptive perspective}},\ }\href@noop {} {\bibfield  {journal} {\bibinfo  {journal} {Psychological Review}\ }\textbf {\bibinfo {volume} {96}},\ \bibinfo {pages} {703} (\bibinfo {year} {1989})}\BibitemShut {NoStop}%
\bibitem [{\citenamefont {Wu}\ \emph {et~al.}(2023)\citenamefont {Wu}, \citenamefont {Bansal}, \citenamefont {Zhang}, \citenamefont {Wu}, \citenamefont {Li}, \citenamefont {Zhu}, \citenamefont {Jiang}, \citenamefont {Zhang}, \citenamefont {Zhang}, \citenamefont {Liu}, \citenamefont {Awadallah}, \citenamefont {White}, \citenamefont {Burger},\ and\ \citenamefont {Wang}}]{autogen2023}%
  \BibitemOpen
  \bibfield  {author} {\bibinfo {author} {\bibfnamefont {Q.}~\bibnamefont {Wu}}, \bibinfo {author} {\bibfnamefont {G.}~\bibnamefont {Bansal}}, \bibinfo {author} {\bibfnamefont {J.}~\bibnamefont {Zhang}}, \bibinfo {author} {\bibfnamefont {Y.}~\bibnamefont {Wu}}, \bibinfo {author} {\bibfnamefont {B.}~\bibnamefont {Li}}, \bibinfo {author} {\bibfnamefont {E.}~\bibnamefont {Zhu}}, \bibinfo {author} {\bibfnamefont {L.}~\bibnamefont {Jiang}}, \bibinfo {author} {\bibfnamefont {X.}~\bibnamefont {Zhang}}, \bibinfo {author} {\bibfnamefont {S.}~\bibnamefont {Zhang}}, \bibinfo {author} {\bibfnamefont {J.}~\bibnamefont {Liu}}, \bibinfo {author} {\bibfnamefont {A.~H.}\ \bibnamefont {Awadallah}}, \bibinfo {author} {\bibfnamefont {R.~W.}\ \bibnamefont {White}}, \bibinfo {author} {\bibfnamefont {D.}~\bibnamefont {Burger}},\ and\ \bibinfo {author} {\bibfnamefont {C.}~\bibnamefont {Wang}},\ }\bibfield  {title} {\bibinfo {title} {{AutoGen}: Enabling next-gen large language model applications via multi-agent conversation},\ }\href
  {https://arxiv.org/abs/2308.08155} {\bibfield  {journal} {\bibinfo  {journal} {arXiv preprint arXiv:2308.08155}\ } (\bibinfo {year} {2023})}\BibitemShut {NoStop}%
\bibitem [{\citenamefont {{LlamaIndex Contributors}}(2025)}]{llamaindex2025}%
  \BibitemOpen
  \bibfield  {author} {\bibinfo {author} {\bibnamefont {{LlamaIndex Contributors}}},\ }\href {https://docs.llamaindex.ai/en/stable/understanding/agent/multi_agent/} {\emph {\bibinfo {title} {Multi-Agent Workflows — LlamaIndex Documentation}}} (\bibinfo {year} {2025})\BibitemShut {NoStop}%
\bibitem [{\citenamefont {Herzig}(2024)}]{sapAI2024}%
  \BibitemOpen
  \bibfield  {author} {\bibinfo {author} {\bibfnamefont {P.}~\bibnamefont {Herzig}},\ }\href {https://news.sap.com/2024/06/ai-innovations-partnerships-sap-sapphire/} {\bibinfo {title} {Sap’s new ai innovations and partnerships deliver real-world results}} (\bibinfo {year} {2024})\BibitemShut {NoStop}%
\bibitem [{\citenamefont {Srivastava}\ \emph {et~al.}(2025)\citenamefont {Srivastava}, \citenamefont {Schuurmans}, \citenamefont {Ray}, \citenamefont {Mesnage}, \citenamefont {Butler}, \citenamefont {Dutta}, \citenamefont {Kubit},\ and\ \citenamefont {Janner}}]{bcgERP2025}%
  \BibitemOpen
  \bibfield  {author} {\bibinfo {author} {\bibfnamefont {J.}~\bibnamefont {Srivastava}}, \bibinfo {author} {\bibfnamefont {J.}~\bibnamefont {Schuurmans}}, \bibinfo {author} {\bibfnamefont {K.}~\bibnamefont {Ray}}, \bibinfo {author} {\bibfnamefont {L.}~\bibnamefont {Mesnage}}, \bibinfo {author} {\bibfnamefont {N.}~\bibnamefont {Butler}}, \bibinfo {author} {\bibfnamefont {S.}~\bibnamefont {Dutta}}, \bibinfo {author} {\bibfnamefont {T.}~\bibnamefont {Kubit}},\ and\ \bibinfo {author} {\bibfnamefont {T.}~\bibnamefont {Janner}},\ }\bibfield  {title} {\bibinfo {title} {Genai can revolutionize erp transformations},\ }\href {https://www.bcg.com/gen-ai-can-revolutionize-erp-transformations} {\bibfield  {journal} {\bibinfo  {journal} {Boston Consulting Group Insight}\ } (\bibinfo {year} {2025})}\BibitemShut {NoStop}%
\bibitem [{\citenamefont {Quach}\ \emph {et~al.}(2023)\citenamefont {Quach}, \citenamefont {Fisch}, \citenamefont {Schuster}, \citenamefont {Yala}, \citenamefont {Sohn}, \citenamefont {Jaakkola},\ and\ \citenamefont {Barzilay}}]{conformalLM2023}%
  \BibitemOpen
  \bibfield  {author} {\bibinfo {author} {\bibfnamefont {V.}~\bibnamefont {Quach}}, \bibinfo {author} {\bibfnamefont {A.}~\bibnamefont {Fisch}}, \bibinfo {author} {\bibfnamefont {T.}~\bibnamefont {Schuster}}, \bibinfo {author} {\bibfnamefont {A.}~\bibnamefont {Yala}}, \bibinfo {author} {\bibfnamefont {J.~H.}\ \bibnamefont {Sohn}}, \bibinfo {author} {\bibfnamefont {T.~S.}\ \bibnamefont {Jaakkola}},\ and\ \bibinfo {author} {\bibfnamefont {R.}~\bibnamefont {Barzilay}},\ }\bibfield  {title} {\bibinfo {title} {Conformal language modeling},\ }\href {https://arxiv.org/abs/2306.10193} {\bibfield  {journal} {\bibinfo  {journal} {arXiv preprint arXiv:2306.10193}\ } (\bibinfo {year} {2023})}\BibitemShut {NoStop}%
\bibitem [{\citenamefont {Li}\ \emph {et~al.}(2025)\citenamefont {Li}, \citenamefont {Wu}, \citenamefont {Wang}, \citenamefont {Xu}, \citenamefont {Hunt},\ and\ \citenamefont {Stein}}]{hmcf2025}%
  \BibitemOpen
  \bibfield  {author} {\bibinfo {author} {\bibfnamefont {Z.}~\bibnamefont {Li}}, \bibinfo {author} {\bibfnamefont {W.}~\bibnamefont {Wu}}, \bibinfo {author} {\bibfnamefont {Y.}~\bibnamefont {Wang}}, \bibinfo {author} {\bibfnamefont {Y.}~\bibnamefont {Xu}}, \bibinfo {author} {\bibfnamefont {W.}~\bibnamefont {Hunt}},\ and\ \bibinfo {author} {\bibfnamefont {S.}~\bibnamefont {Stein}},\ }\bibfield  {title} {\bibinfo {title} {{HMCF}: A human-in-the-loop multi-robot collaboration framework based on large language models},\ }\href {https://arxiv.org/abs/2505.00820} {\bibfield  {journal} {\bibinfo  {journal} {arXiv preprint arXiv:2505.00820}\ } (\bibinfo {year} {2025})}\BibitemShut {NoStop}%
\bibitem [{\citenamefont {Naminas}(2025)}]{labelyourdata2025}%
  \BibitemOpen
  \bibfield  {author} {\bibinfo {author} {\bibfnamefont {K.}~\bibnamefont {Naminas}},\ }\href {https://labelyourdata.com/articles/human-in-the-loop-in-machine-learning} {\bibinfo {title} {Human in the loop machine learning: The key to better models}} (\bibinfo {year} {2025})\BibitemShut {NoStop}%
\end{thebibliography}%

\end{document}